\documentclass[letterpaper]{article} % DO NOT CHANGE THIS
\usepackage{aaai25}  % DO NOT CHANGE THIS
\usepackage{times}  % DO NOT CHANGE THIS
\usepackage{helvet}  % DO NOT CHANGE THIS
\usepackage{courier}  % DO NOT CHANGE THIS
\usepackage[hyphens]{url}  % DO NOT CHANGE THIS
\usepackage{graphicx} % DO NOT CHANGE THIS
\usepackage{natbib}  % DO NOT CHANGE THIS AND DO NOT ADD ANY OPTIONS TO IT
\usepackage{caption} % DO NOT CHANGE THIS AND DO NOT ADD ANY OPTIONS TO IT
\usepackage{multirow}
\usepackage{multicol}
\usepackage{booktabs}
\usepackage{algorithm}
\usepackage{algorithmic}
\usepackage{graphicx} % DO NOT CHANGE THIS
\usepackage{multirow}
\usepackage{makecell}
\usepackage{enumitem}
\usepackage{amsmath}

\usepackage{booktabs}
\usepackage{amsmath}
\usepackage{multirow}
\usepackage{bbding}
\usepackage{mathrsfs}
\usepackage{subfigure}
\usepackage{amssymb}
\usepackage{mathtools}
\usepackage{xcolor}
\usepackage[most]{tcolorbox}
\usepackage{hyperref}
\usepackage{newfloat}
\usepackage{listings}
\DeclareCaptionStyle{ruled}{labelfont=normalfont,labelsep=colon,strut=off} % DO NOT CHANGE THIS
\floatstyle{ruled}
\newfloat{listing}{tb}{lst}{}
\floatname{listing}{Listing}
\title{Toward General Instruction-Following Alignment for Retrieval-Augmented Generation}
\author{
    Guanting Dong\textsuperscript{\rm 1},
    Xiaoshuai Song\textsuperscript{\rm 2},
    Yutao Zhu\textsuperscript{\rm 1},
    Runqi Qiao\textsuperscript{\rm 2},
    Zhicheng Dou\textsuperscript{\rm 1}\thanks{Corresponding author},
    Ji-Rong Wen\textsuperscript{\rm 1}
}
\affiliations{
    \textsuperscript{\rm 1}Gaoling School of Artificial Intelligence, Renmin University of China.\\
    \textsuperscript{\rm 2}School of Artificial Intelligence, Beijing University of Posts and Telecommunications \\
    \small \texttt{\{dongguanting,dou\}@ruc.edu.cn} \\

}

\usepackage{bibentry}
\begin{document}

\maketitle

\begin{abstract}
Following natural instructions is crucial for the effective application of Retrieval-Augmented Generation (RAG) systems. Despite recent advancements in Large Language Models (LLMs), research on assessing and improving instruction-following (IF) alignment within the RAG domain remains limited. To address this issue, we propose VIF-RAG, the first automated, scalable, and verifiable synthetic pipeline for instruction-following alignment in RAG systems. We start by manually crafting a minimal set of atomic instructions ($<$100) and developing combination rules to synthesize and verify complex instructions for a seed set. We then use supervised models for instruction rewriting while simultaneously generating code to automate the verification of instruction quality via a Python executor. Finally, we integrate these instructions with extensive RAG and general data samples, scaling up to a high-quality VIF-RAG-QA dataset ($>$100k) through automated processes.
To further bridge the gap in instruction-following auto-evaluation for RAG systems, we introduce FollowRAG Benchmark, which includes approximately 3K test samples, covering 22 categories of general instruction constraints and four knowledge-intensive QA datasets. Due to its robust pipeline design, FollowRAG can seamlessly integrate with different RAG benchmarks. Using FollowRAG and eight widely-used IF and foundational abilities benchmarks for LLMs, we demonstrate that VIF-RAG markedly enhances LLM performance across a broad range of general instruction constraints while effectively leveraging its capabilities in RAG scenarios. Further analysis offers practical insights for achieving IF alignment in RAG systems. Our code and datasets are released at \url{https://FollowRAG.github.io}.

\end{abstract}


\section{1. Introduction}

The advancement of Large Language Models (LLMs)~\citep{openai2024gpt4,yang2024qwen2} has profoundly revolutionized a variety of real-world tasks expressed in natural language~\citep{wei2023chainofthought,luo2023wizardmath}. However, they still suffer from hallucinations and factual inconsistencies~\citep{bang2023multitask}, impacting the authenticity of generated answers. Retrieval-Augmented Generation (RAG) has gained recognition as a promising solution, empowering LLMs to leverage reliable information from retrieved documents, thereby returning high-quality responses~\citep{guu2020realm,lewis2021retrievalaugmented}.

% In real-world interaction scenarios, users often deviate from standard templates when posing questions, instead imposing diverse instructions on model outputs to meet specific task requirements~\citep{zhang2023instruction,chung2024scaling}. Consequently, enhancing instruction-following (IF) capabilities is foundational to the broad application of LLM and RAG systems, with the core goal of enabling models to adapt to the diverse intents of users, which has already garnered widespread attention in the AI field.

\begin{figure}[t]
    \centering
    \includegraphics[width=0.85\linewidth]{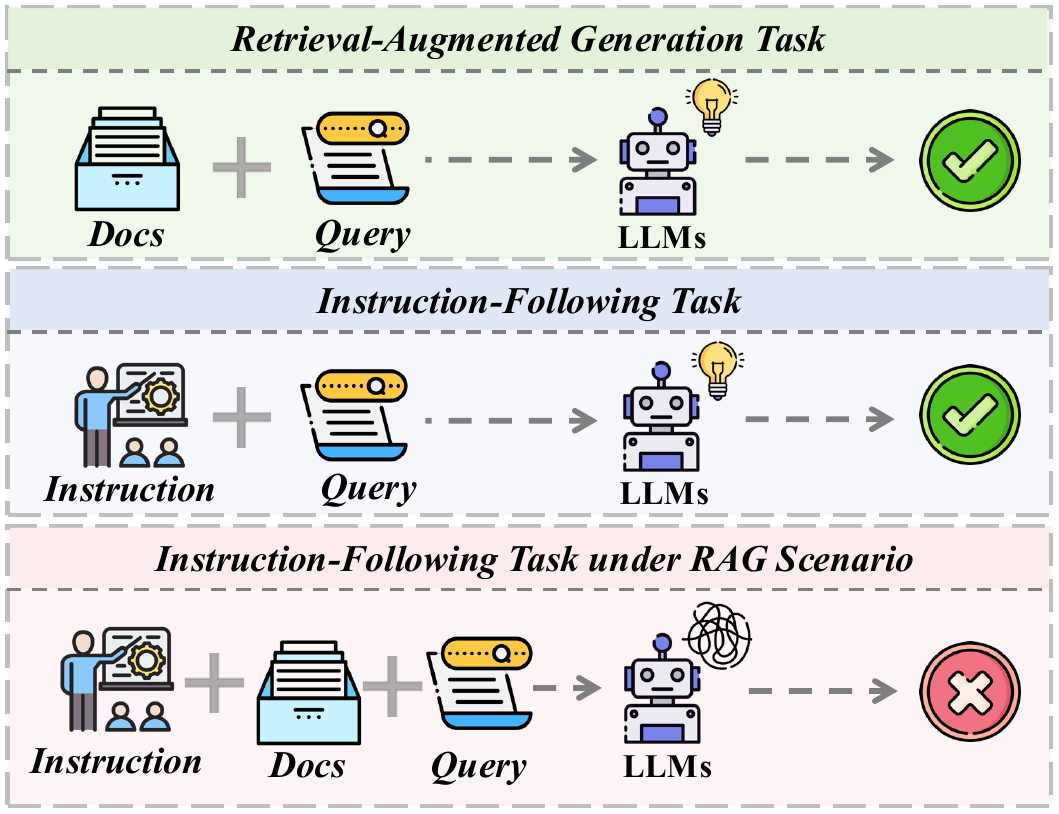}
     \vspace{-0.5em}
    \caption{The task format of instruction-following tasks for LLMs in RAG scenarios.}
   
    \label{fig:intro1}
    \vspace{-1em}
\end{figure}
In real-world interaction scenarios, users often deviate from standard templates when posing questions, instead of imposing diverse instructions on model outputs to meet specific task requirements~\citep{jiang2023followbench,chung2024scaling}. Consequently, improving instruction-following (IF) capabilities is foundational to the effective application of LLM and RAG systems. The core goal of IF is to enable models to adapt to the diverse intents of users, which has garnered widespread attention in the LLM community.

\begin{figure*}[t]
    \centering
    \includegraphics[width=\linewidth]{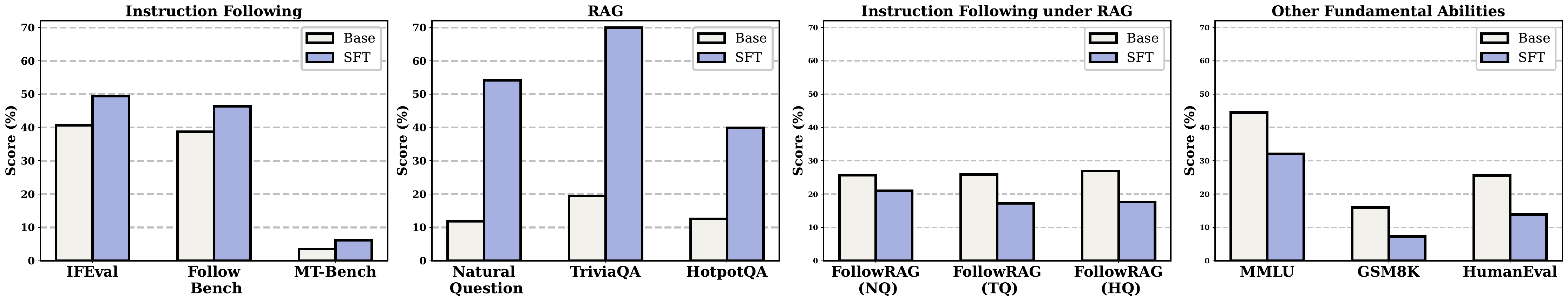}
     \vspace{-2em}
    \caption{The performance comparison between Mistral-7B base and SFT version models on different tasks. The SFT version refers to the base model that has been fine-tuned using a mixed dataset, including NQ, TQ, HQ, and ShareGPT. Details results and setups can be found in Table \ref{tab:1} \& \ref{table2}}
   
    \label{fig:intro2}
    \vspace{-1em}
\end{figure*}

% Existing efforts on instruction-following alignment primarily focuses on multi-grained evaluation~\citep{zhou2023instructionfollowing,jiang2024followbench,qin2024infobench,xia2024fofo,yan2024refutebench,wen2024benchmarking} and high quality instruction data synthesis~\citep{wang2024codeclm,sun2024conifer,zhao2024self} to enhance LLM's  natural instruction-following capabilities. However, when faced with more complex RAG scenarios, the diverse knowledge introduced by retrieval-augmented techniques poses significant challenges for LLMs in handling complex instructions. As shown in figure x, after supervised fine-tuning on high-quality RAG and general capabilities datasets, the LLM's individual instruction-following and RAG abilities both improved significantly. However, in RAG scenarios, the model struggles to generalize by effectively combining these two capabilities, sometimes leading to performance conflicts. Unfortunately, research on instruction following within RAG systems remains relatively underdeveloped, which significantly hinders their practical applications~\citep{petroni2019language,pmlr-v133-min21a,dong2023abilities,zhu2024one}.

Existing efforts on instruction-following alignment primarily focus on multi-grained evaluation~\citep{zhou2023instructionfollowing,jiang2024followbench,wen2024benchmarking} and high-quality instruction data synthesis~\citep{sun2024conifer,zhao2024self} to enhance LLMs' natural instruction-following capabilities. However, in complex RAG scenarios, the diverse knowledge introduced by retrieval-augmented techniques presents significant challenges for LLMs in effectively handling complex instructions (Figure \ref{fig:intro1}). As shown in Figure \ref{fig:intro2}, after supervised fine-tuning on high-quality general and knowledge-intensive QA datasets, LLMs demonstrate robust performance in both IF and RAG tasks (Mistral-base vs. Mistral-SFT). However, these capabilities do not always generalize well to instruction-following tasks under RAG scenarios and may even conflict with the performance of other fundamental abilities~\citep{dong2023abilities,zhu2024one}. Unfortunately, research on instruction-following in RAG systems remains limited, significantly hindering their application in real-world interactions.

To tackle these challenges, our aim is to address following critical research questions:

\begin{itemize}
\item \textit{\textbf{RQ1.} How can we comprehensively evaluate the complex instruction-following capabilities in the RAG scenario?}
\item \textit{\textbf{RQ2.} How can we achieve scalable and reliable instruction-following alignment in RAG systems while preserving the it's foundational abilities from conflict?}
\end{itemize}
In this paper, we propose VIF-RAG, the first automated, scalable, and reliable data synthesis pipeline for achieving complex instruction-following alignment in RAG scenarios. The core insight of VIF-RAG is to ensure every step of data augmentation and combination includes a proper verification process. Specifically, we start by manually crafting a minimal set of atomic instructions ($<$100) and developing combination rules to synthesize and verify complex instructions for a seed set. We then use supervised models for instruction rewriting. Motivated by tool execution studies~\citep{le2022coderl,qiao2024making}, we employ the same supervised model to generate verification code and automatically verify the quality of augmented instructions through the Python compiler's outputs. Finally, we combine these high-quality instructions with RAG datasets from various domains (each containing retrieved documents per query), performing the augmentation and dual validation process to synthesize a high-quality instruction-based RAG dataset, named VIF-RAG-QA ($>$100K samples).

To further bridge the gap in automatic instruction-following evaluation for RAG systems, we introduce FollowRAG, the first benchmark dedicated to comprehensively assessing the complex instruction-following capabilities of RAG systems. FollowRAG aggregates constraints from real-world scenarios. It includes approximately 3K test samples, spanning 4 knowledge-intensive QA benchmarks and 22 types of constraints. Due to its robust pipeline design, FollowRAG can seamlessly integrate with different RAG benchmarks.

To summarize, our contributions are as follows:

\begin{itemize}

\item To first achieve instruction-following alignment in the RAG system, we propose VIF-RAG, the first automated, scalable, and verifiable data synthetic framework. VIF-RAG uniquely combines augmented rewriting with diverse validation processes to synthesize high-quality instruction-following alignment data from almost scratch ($<$100), scaling up to over 100K samples.

\item We introduce FollowRAG, the first benchmark designed to comprehensively evaluate LLM's complex instruction-following abilities in RAG tasks. FollowRAG includes nearly 3K test samples, spanning four knowledge-intensive QA benchmarks and 22 types of constraints. Its design ensures seamless integration with various RAG benchmarks, providing strong scalability.

% We introduce FollowRAG, the first benchmark for evaluating LLMs' complex instruction-following abilities in RAG tasks. FollowRAG includes over 3,000 test samples, spanning four knowledge-intensive QA benchmarks and 22 types of constraints. Its design ensures seamless integration with various RAG benchmarks, providing strong scalability.

% \item With FollowRAG and evaluating 4 widely-used IF metrics and 4 foundational LLM benchmarks, we demonstrate that different open-source LLMs with our VIF-RAG-QA synthetic dataset achieve great performance across general instruction constraints in both RAG and standard scenarios while effectively preserving other foundational capabilities. Further
% analysis offers practical insights for optimizing IF alignment in RAG systems.

\item With FollowRAG and 8 widely-used IF and 3 foundational abilities benchmarks, we demonstrate that different LLMs with VIF-RAG achieve extraordinary alignment on general instruction following in both RAG and standard scenarios while effectively preserving other foundational capabilities. Further
analysis offers practical insights for optimizing IF alignment in RAG systems.

\end{itemize}

\section{2. Related Work}
\textbf{Instruction-Following Alignment for LLMs.}
Instruction-following ability is a core capability of large language models. Existing works fall into two main categories. The first includes efforts like MMLU and MTbench~\citep{hendrycks2021measuring,zheng2024judging}, which rigorously evaluate models' adherence to general instructions. Moreover, works like IFEval and Followbench~\citep{zhou2023instructionfollowing,jiang2024followbench} focus on fine-grained assessment under specific constraints, using stricter criteria such as instruction difficulty, domain, and task formats~\citep{qin2024infobench,xia2024fofo,yan2024refutebench,wen2024benchmarking}. The other category focuses on improving IF alignment. Manual design of instructions and responses by human annotators~\citep{wei2021finetuned} is challenging and costly. To address this, methods are developed to synthesize diverse instructions, allowing weaker models to mimic the responses of advanced models~\citep{dubois2024alpacafarm,dong2024self,xu2023wizardlm}, achieving strong-to-weak alignment~\citep{cao2024towards}.

\textbf{Alignment for Retrieval-Augmented Generation.} Retrieval-Augmented Generation (RAG) addresses the issue of knowledge hallucination in LLMs by retrieving relevant factual information, offering a promising solution~\citep{guu2020realm,lewis2021retrievalaugmented}. However, efficiently aligning retrieved knowledge with LLMs' preferences remains a challenge. Researchers have developed robust reranker-based methods~\citep{sun2023chatgpt,qin2024large,ma2023zero} and data filtering approaches~\citep{wang2023learning} to reduce noisy information and bridge this gap. Additionally, approaches like RePLUG~\citep{shi2023replug} integrate LLMs' preferences into training objectives to improve alignment. Query rewriting methods~\citep{ma2023query,ren2023investigating} attempt to adjust inputs based on these preferences. Furthermore, SelfRAG and MetaRAG~\citep{asai2024selfrag,zhou2024metacognitive} use multi-round retrieval and generation to refine outputs and achieve better alignment. Despite these advancements, the diverse knowledge introduced by retrieval-augmented techniques poses significant challenges for LLMs in handling complex instructions. This highlights the need for further exploration into achieving effective instruction-following alignment in RAG systems.

% Retrieval-Augmented Generation aims to alleviate the knowledge hallucination in LLMs by retrieving relevant factual knowledge, providing a promising solution~\citep{guu2020realm,lewis2021retrievalaugmented}. However, due to the inherent gap between the retrieved knowledge and the knowledge preferences of LLM reader, efficiently aligning the knowledge preferences within RAG systems has become a sustained focus for researchers. A series of robust reranker-based methods~\citep{sun2023chatgpt,qin2024large,ma2023zero} and data filtering approach~\citep{wang2023learning} have emerged to efficiently eliminate noisy knowledge to bridge the knowledge gap. 

\begin{figure*}[t]
    \centering
    \includegraphics[width=0.93\linewidth]{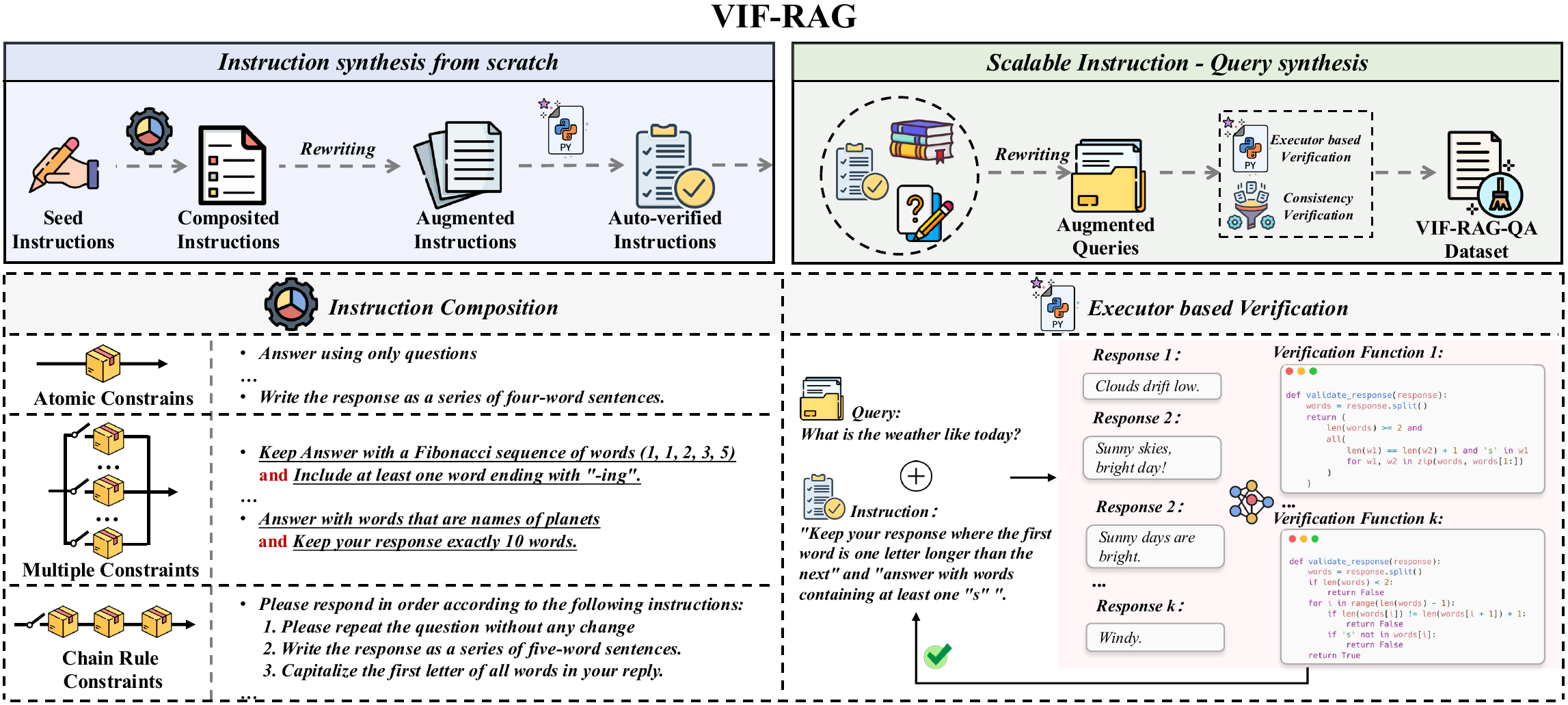}
    % \vspace{-1em}
    \vspace{-0.5em}
    \caption{The overall framework of VIF-RAG. The top section presents the pipeline for automated IF data synthesis in RAG scenario, while the bottom section shows examples of 'Instruction Composition' and 'Executor-based Verification,' respectively.}
    \label{fig:main}
    \vspace{-1em}
\end{figure*}

% Furthermore, several works such as RePLUG~\citep{shi2023replug}, KnowPAT~\citep{zhang2024knowledgeable} incorporate LLM's preference information into training objectives, enabling the models to understand the aligned knowledge. Query rewriting methods~\citep{ma2023query,ren2023investigating,dong2024understand} aim to directly align by rewriting inputs according to preferences. Additionally, SelfRAG and MetaRAG~\citep{asai2024selfrag,zhou2024metacognitive} employ multi-round retrieval and generation to refine their outputs, achieving knowledge preference alignment. However, while these methods primarily focus on aligning knowledge preferences, the diverse knowledge introduced by retrieval-augmented techniques presents significant challenges for LLMs in handling complex instructions. This more foundational issue allows us to further explore how to achieve instruction-following alignment in RAG systems.

\section{3. Preliminaries}\label{sec:preliminaries}

\textbf{Retrieval-Augmented Generation (RAG).}\label{sec:31} Retrieval-Augmented Generation systems usually operates under a \textit{retrieve-then-read} framework~\citep{lewis2021retrievalaugmented}. The external retriever is integrated to gather supporting knowledge and improve the generation process. Given a query $q$, a retriever $R$ recalls $k$ relevant documents $D_{q}=\{d_i\}^{k}_{i=1}$ from an external corpus comprised of $N$ documents. We employ the DPR~\citep{karpukhin2020dense} to obtain hidden vectors for queries and documents. The relevance score is determined by measuring the dot-product similarity between the query and document representations, allowing the retrieval of the top-$k$ documents $D_{q}$::
\begin{align}
D_{q} = {\text{argtop-}k} \left[ E_{\text{d}} (d_i)^{\top} \cdot E_{\text{q}}(q) \mid i = \{1 \ldots N\} \right].
\end{align}

Then, the retrieved documents are concatenated with the query into an LLM reader $R$ to generate the target text: 
\begin{align}
    y = R(q,D_{q}) = {\log{{P}_{\theta}\left(q, D_{q}\right)}},
\label{eq::task}
\end{align}
where ${P}_{\theta}$ is the output probability distribution. 

\textbf{Instruction-following Alignment for RAG.} Following instructions is one of the most foundational ability for LLMs in RAG systems. Given an instruction $I = {\{I_{j}\}}_{j=1}^{M}$ with $M$ specific constraints and a specific query $q$ with corresponding relevant $k$ retrieved documents $D_{q}$, The LLM $\pi_{\theta}$ in the RAG system is expected to produce an accurate response $y \sim \pi_{\theta}(y \mid q, D_q, I)$ while obeying with the specified constraints.

\section{4. VIF-RAG Framework}
In this section, we propose VIF-RAG, a verifiable automated instruction data synthesis framework for RAG scenarios. The core design of VIF-RAG is that each step of the automated generation or combination is accompanied by an appropriate verification process. ViF-RAG framework can be broadly split into two sections: (1) the instruction synthesis stage and (2) instruction-query synthesis, scaling from almost scratch ($<$100) to over 100K high-quality instruction-query samples. Below, we will delve into the specifics.

\subsection{4.1. Instruction Synthesis from Scratch}

\textbf{Handwritten Seed Instructions.} We initially develop a minimal seed instruction set $D_{\text{seed}}^{\text{atom}}$ manually, using four foundational categories of constraints: \textit{format constraints, semantic constraints, knowledge constraints, and lexical constraints}, as themes for instruction writing. The following presents specific criteria regarding the 4 constraints:

\begin{itemize}[leftmargin=1em]
\item \textbf{Format Constraints} require the output to adhere to specific standards in terms of format, length, and structure. The content should be organized, clear, and meet the required format specifications.
\item \textbf{Semantic Constraints} require the output's theme, language style, personality, and sentiment to align with the given instructions. The content should be semantically consistent with expectations and adhere to the specified tone or expression.
\item \textbf{Knowledge Constraints} require the output to be accurate, comprehensive, and in-depth. The content should be informative, cover all necessary information, and maintain consistency in knowledge expression.
\item \textbf{Lexical Constraints} require the output to include specific keywords or phrases, ensuring precision and relevance in word choice. The content should meet the expected requirements in terms of vocabulary selection.

\end{itemize}

We hire only one well-educated human annotator to manually create 15 single-atomic instructions for each type of constraint. Notably, this is the only process in our data synthesis process that includes human supervision. 

% Detailed samples of the $D_{\text{seed}}^{\text{atom}}$ and the principles of the constraints are listed in the Appendix.

\textbf{Instruction Composition \& Verification.} Real-world instructions often involve multiple constraints in one user query. To address this complexity, we design rules to automatically combine atomic instructions into diverse, complex instructions:

\begin{itemize}
\item \textit{\textbf{Multiple Constraints}}: As illustrated in Figure \ref{fig:main}, we randomly sample pairs of instructions from $D_{\text{seed}}^{\text{atom}}$ and insert them into a constraint template. By directly concatenating these instruction pairs, we create complex instructions that contain dual and triple constraints. This type of instruction requires the model to generate results that satisfy multiple constraints simultaneously.

\item \textit{\textbf{Chain Rule Constraints}}: We design sequential conditional constraint templates and selected atomic instructions from $D_{\text{seed}}^{\text{atom}}$ to form chain constraints. Formally, the chain consists of $n$ tasks $\{T_1, T_2, . . . , T_n\}$, requiring the model's output to complete these n tasks sequentially.
\end{itemize}

\textbf{Verification.} Randomly combining these atomic instructions can easily lead to conflicts between them (e.g., don't use words containing the letter 'I', use words that end with '-ing'). These semantic conflicts can be challenging to detect using a simple Natural Language Inference model. To detect potential conflicts between these instructions, we use a robust supervised model that rates their consistency from 1 to 10. Samples scoring below 8 are excluded to refine our high-quality complex instruction set $D_{\text{seed}}^{\text{complex}}$.  Ultimately, we arrive at the initial seed instruction set $D_{\text{seed}} = \{D_{\text{seed}}^{\text{atom}} \cup D_{\text{seed}}^{\text{complex}}\}$. Detailed information about the prompt templates are listed in the Appendix.

\textbf{Instruction Rewriting \& Quality Verification.} To automate the scaling up of instructions, the instruction rewriting strategy is considered the most natural augmentation method, and has received significant attention in the RAG and reasoning fields~\citep{mumuni2022data,xie2020unsupervised,rft,mugglemath,dotamath}. We use a supervised model\footnote{For the supervised model, we use GPT-4-turbo-2024-04-09. We conduct the ablation for supervision model in Table \ref{tab:ablation_sup}.} to iteratively rewrite instructions from the $D_{\text{seed}}$ set in batches of 50 for $K$ rounds, generating an augmented set $D_{\text{aug}}$. Subsequently, we merge the seed and augmented samples to form the combined instruction set $D_{\text{ins}} = {D_{\text{seed}} \cup D_{\text{aug}}}$, removing duplicates.

Inspired by tool execution works~\citep{le2022coderl}, we aim to leverage the powerful coding abilities of LLMs to assist in verifying the quality of auto-generated instructions. As shown in Figure \ref{fig:main}, for each instruction $I \in D_{\text{ins}}$, we use the supervision model to generate $K$ verification function codes and corresponding test cases $\{func_{j}^I, c_{j}^I\}_{j=1}^{K} \in D_{\text{verify}}$, and assess the instruction's quality by analyzing the output of the executor $\mathcal{E}$. For any function and test case $\{func_j^I, c_j^I\} \in D_{\text{verify}}$, its execution output is:
\begin{equation}
% \resizebox{1\hsize}{!}{$
\mathcal{E}(func_j^I, c_j^I)=\begin{cases} 1 & \text{If output is ``True''} \\0& \text{If output is ``False'' or ``Error''} \end{cases} 
% $}
\end{equation}
Therefore, we can calculate the accuracy $Acc_{\text{func}}$ of each verification function based on $K$ test samples, as well as the accuracy $Acc_{\text{case}}$ of each case evaluated using $K$ verification functions. These can be formulated as:

\begin{equation}
\begin{cases}
Acc_{\text{func}}=\frac{1}{K} \sum_{j=1}^{K}  \mathcal{E}(func^I, c_j^I)_{j=1}^{K}
\\
\\
Acc_{\text{case}}=\frac{1}{K} \sum_{j=1}^{K}  \mathcal{E}(func_j^I, c^I)_{j=1}^{K}
\end{cases} 
\end{equation}
Based on the above cross metrics, we require that at least one \( Acc_{\text{func}} \) and \( Acc_{\text{case}} \) of the each instruction must exceed 0.5, Ultimately, we obtain the auto-verified instruction set as 
\begin{equation}
\small
D_{\text{ins}}^{\text{verify}} = \{ d \in D_{\text{ins}} \mid Acc_{\text{func}}(d) > 0.5 \ \& \ Acc_{\text{case}}(d) > 0.5 \}
\end{equation}
The samples that do not meet the cross metrics are discarded.

% We retained fifi if Acc_{fi} > 0.5; if both (Acc_{fi}, Acc_{t

\subsection{4.2. Scalable Instruction-Query Synthesis} 

\textbf{Random Instruction-Query Combination.} In real-world interactions with RAG systems, achieving IF alignment depends on effectively integrating the synthesized instructions with the queries used by the RAG system. To meet this goal, as depicted in Figure \ref{fig:main}, we first extract high-quality queries from two different data sources.

\textit{\textbf{1) RAG Domain:}} 
Building effective RAG system need to prepare sufficient amounts of QA-format data with relevant knowledge to enhance its knowledge-based interaction capabilities. Consequently, we randomly select a query set $Q$ from mixed QA data sources, including open-domain multi-hop and knowledge base QA scenarios \footnote{We use the training sets from Natural Questions, TriviaQA, HotpotQA, and WebQuestionsSP as mixed QA sources.}. Following the \textit{retrieve-then-read} paradigm~\citep{lewis2021retrievalaugmented}, We employ the dense retriever $R$ to fetch the top-$K$ relevant documents $D_i$ for each query $q \in Q$ from an external knowledge base, resulting in the dataset $D_{\text{RAG}} = \{q_i, D_i\}_{i=1}^K$. Furthermore, we randomly select $K$ queries along with their corresponding retrieved documents from $D_{\text{RAG}}$ for each instruction $I$ and combine them to create the RAG query set with IF constraints $D_{\text{IF-RAG}} = \{I_j, q_j, D_j\}_{j=1}^{K}$.

\begin{figure*}[t]
    \centering
    \includegraphics[width=0.9\linewidth]{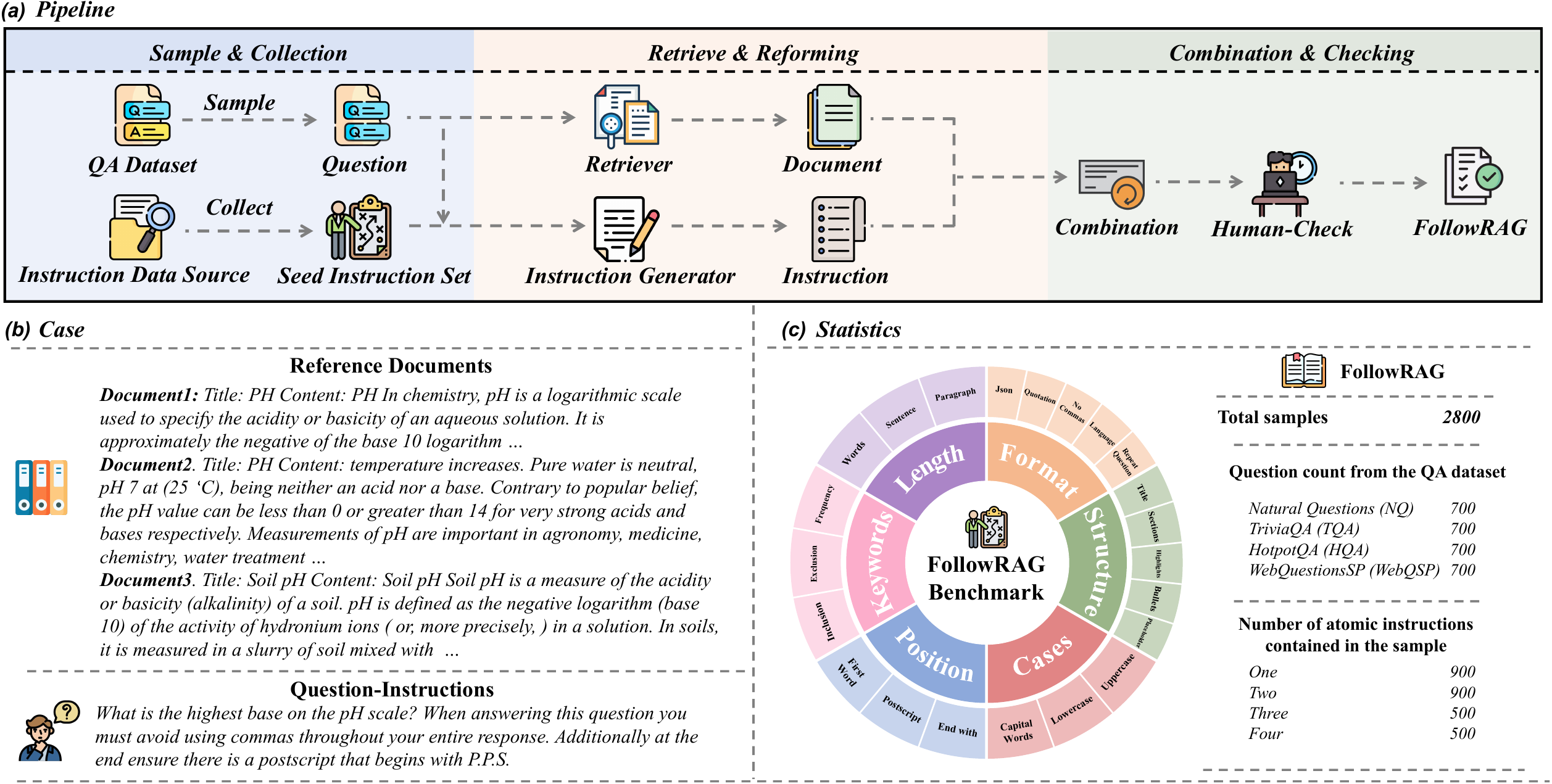}
    \vspace{-0.5em}
    \caption{The construction pipeline, diagram and statistics of FollowRAG.}
    \label{fig:bench}
    \vspace{-1em}
\end{figure*}

\textit{\textbf{2) General Domain:}} In addition to incorporating RAG-specific abilities, the RAG system has to possess basic human-aligned abilities to meet users' daily interaction needs. Therefore, ShareGPT~\citep{vicuna2023}, which provides authentic multi-turn human dialogue data, is our natural choice. Similar to how we handle the RAG domain, for each instruction $I \in D_{\text{ins}}$, we randomly select $K$ queries from the ShareGPT to combine with the instruction and construct the general dataset $D_{\text{IF-General}}$ for each instruction.

Ultimately, we merge the instruction-constrained query sets from these two domains into the final query set of VIF-RAG-QA, formulated as $D_{\text{VIF-RAG}}^{q}$.

\textbf{Instruction-Query Rejection Sampling.} It is worth noting that under diverse instruction-following constraints, the original grounding truth answers for queries in both the RAG and general datasets become unreliable. To address this issue and improve synthetic data diversity, we adopt a rejection sampling strategy~\citep{yuan2023scaling}. Specifically, we use the supervision model to generate $K$ responses $y_x = \{y_i\}_{i=1}^K$ for each instruction-query pair $x \in D_{\text{VIF-RAG}}^q$, resulting in $\{x, y_x\} \in D_{\text{VIF-RAG}}$.

\textbf{Dual Stage Verification.}
To further ensure comprehensive quality control of the synthetic dataset, we employ a dual stage verification process for the instruction-query data:

\begin{itemize}
\item \textit{\textbf{Executor-based Verification:}} To automatically verify whether model-generated responses comply with the constraints of the instruction-query samples, we leverage pre-existing verification functions to evaluate adherence in the augmented outputs. As in the ``Instruction Rewriting \& Quality Verification'' section, at least one response in $D_{\text{VIF-RAG}}$ must achieve an accuracy rate $Acc_{\text{case}}$ above 0.5 across all verification functions; otherwise, the sample is discarded.

\item \textit{\textbf{Consistency Verification:}} We have noticed that combined instructions and queries often conflict. A simple example is when the query “Please write a brief biography of Barack Obama.” does not meet the instruction “Strictly limit your answer to less than 10 tokens.” Building on previous consistency verification of instructions, we employ a supervision model to evaluate the alignment between queries and instructions on a scale of 1 to 10, discarding samples that receive a score below 8.
\end{itemize}

After dual stage verification, we have automatically obtained a large-scale, high-quality VIF-RAG-QA dataset.

\section{5. FollowRAG Benchmark}

To bridge the gap in automatic instruction-following evaluation for RAG systems, we introduce FollowRAG in this section. We provide a detailed introduction from two aspects: ``Data Construction'' and ``Evaluation and Statistics''.

\subsection{5.1. Dataset Construction}
\textbf{Instruction Collection \& Extraction.} FollowRAG aims to assess the model's ability to follow user instructions in complex multi-document contexts. Drawing from general IF datasets like IFEval~\citep{ifeval} and FollowBench~\citep{followbench}, we collect and verify definitions and examples of atomic instructions using rules (e.g., code), excluding those irrelevant to RAG scenarios. Ultimately, we identify 22 types of instruction constraints, encompassing language, length, structure, and keywords.

\textbf{Instruction Reforming.} We use widely-used question-answering (QA) datasets, such as Natural Questions ~\citep{NQ}, as the foundation for constructing FollowRAG samples. For a query sampled from the QA datasets, we need to generate a complex instruction containing $n$ atomic instruction constraints (with $n$ ranging from 1 to 4). To enhance the diversity of atomic instruction representations, we employ GPT-4o as the instruction generator. Specifically, given a query, we first sample $n$ instructions from the atomic instruction set and perform conflict detection. Subsequently, with examples as demonstrations, we prompt the LLM to generate a new varied instruction text for each type of atomic instruction, along with parameters for instruction-following evaluation.

\textbf{Combination.} Finally, we integrate the retrieved passages, query and atomic instructions to construct the sample input for FollowRAG. To avoid mechanically concatenating the query and instructions in a template-based manner, we prompt supervised model to naturally blend the multiple atomic instructions and the query into a coherent instruction-query paragraph. We then add the top-$K$ document passages retrieved based on the query to the instruction-query paragraph, forming the complete sample input.

% Finally, we integrate the retrieved passages, query, and atomic instructions to construct the sample input for FollowRAG. To avoid simply concatenating the query and instructions, we prompt the supervised model to blend them into a coherent instruction-query paragraph. We then add the top-$K$ document passages retrieved based on the query to this paragraph to form the complete sample input.

\subsection{5.2. Evaluation and Statistics}
After obtaining the model's output, we evaluate it from two perspectives: instruction following and question answering (QA) under the RAG paradigm:
\begin{itemize}
\item \textit{\textbf{Instruction Following:}} Utilizing the verifiable nature of our atomic instructions and following the IFEval approach, we automate the verification of the model’s adherence to each instruction through code validation. We then calculate the average pass rate for each atomic instruction across all samples to determine the instruction-following score in FollowRAG.
\item \textit{\textbf{RAG:}} Under new instruction constraints, the model's target output differs from  the gold answers in the original QA dataset, rendering traditional metrics like Exact-Match ineffective. To address this, we use the original gold answers as a reference and utilize GPT-4o to evaluate whether the model's outputs correctly address the questions. The scoring criteria are as follows: Completely correct (1 point), Partially correct (0.5 points), Completely incorrect (0 points). The average score of all samples is taken as the RAG score for FollowRAG.
\end{itemize}

% \subsection{Statistics}
For detailed statisticsin in Figure~\ref{fig:bench}, FollowRAG is the first instruction-following evaluation dataset under  RAG scenario comprising 2.8K samples, covering 22 fine-grained atomic instructions across 6 categories. The queries in FollowRAG are sourced from 4 QA datasets across 3 types: 1) Open-Domain QA: \textbf{Natural Questions (NQ)}~\citep{NQ} and \textbf{TriviaQA (TQA)}~\citep{TriviaQA}; 2) Multi-Hop QA: \textbf{HotpotQA (HQA)}~\citep{HotpotQA}; and 3) Knowledge Base QA: \textbf{WebQuestionsSP (WebQSP)}~\citep{webqsp}. To further construct varying levels of instruction-following difficulty, FollowRAG includes 0.9K samples of single and dual atomic instructions, as well as 0.5K complex multi-instruction samples containing 3 and 4 atomic instructions, respectively.

\section{6. Experiment}

\begin{table*}[!t]
    \centering
    \resizebox{\textwidth}{!}{%
    \begin{tabular}{lccccccccccccccc}
    \toprule
        \multirow{2}{*}{\textbf{Model}} & \multicolumn{3}{c}{\textbf{NQ}} & \multicolumn{3}{c}{\textbf{TQ}} & \multicolumn{3}{c}{\textbf{HQ}} &\multicolumn{3}{c}{\textbf{WebQSP}} & \multicolumn{3}{c}{\textbf{ALL}} \\ 
        \cmidrule(r){2-4}\cmidrule(lr){5-7}\cmidrule(lr){8-10}\cmidrule(lr){11-13}\cmidrule(l){14-16}
         & IF & RAG & AVG & IF & RAG & AVG & IF & RAG & AVG & IF & RAG & AVG & IF & RAG & AVG \\ \midrule
        Llama3-8B-base & 3.2 & 5.7 & 4.4 & 4.1 & 15.9 & 10.0 & 3.6 & 7.3 & 5.5 & 10.0 & 23.1 & 16.5 & 5.2 & 13.0 & 9.1 \\ 
        Llama3-8B-SFT & \underline{15.7} & \underline{59.5} & \underline{37.6} & \underline{15.0} & \underline{76.5} & \underline{45.7} & \underline{15.0} & \textbf{52.5} & \underline{33.8} & \underline{14.4} & \underline{70.0} & \underline{42.2} & \underline{15.0} & \underline{64.6} & \underline{39.8} \\ 
        Llama3-8B-SFT-VIF-RAG & \textbf{43.9} & \textbf{65.0} & \textbf{54.5} & \textbf{42.7} & \textbf{78.0} & \textbf{60.4} & \textbf{39.6} & \underline{46.0} & \textbf{42.8} & \textbf{42.5} & \textbf{70.5} & \textbf{56.5} & \textbf{42.2} & \textbf{64.9} & \textbf{53.5} \\ 
         \midrule
         
        Mistral-7B-base & 25.7 & 31.1 & 28.4 & 25.9 & 44.4 & 35.2 & 26.9 & 19.9 & 23.4 & 24.7 & 20.4 & 22.6 & 25.8 & 29.0 & 27.4 \\ 
        Mistral-7B-SFT & 21.0 & 48.5 & 34.7 & 17.2 & \textbf{71.5} & 44.3 & 17.6 & \textbf{46.5} & 32.1 & 21.7 & \textbf{66.5} & 44.1 & 19.3 & \textbf{58.3} & 38.8 \\ 
        
        Mistral-7B-SFT Conifer & 29.9 & \underline{49.5} & 39.7 & 30.5 & 67.0 & 48.7 & 26.5 & 40.0 & 33.2 & 31.1 & \underline{63.0} & \underline{47.1} & 29.5 & 54.9 & 42.2 \\ 
        Mistral-7B-SFT Evol-Instruct & \underline{41.7} & 41.5 & \underline{41.6} & \underline{37.0} & 63.5 & \underline{50.4} & \underline{35.4} & 35.0 & \underline{35.2} & \underline{39.4} & 54.0 & 46.7 & \underline{38.4} & 48.5 & \underline{43.5}  \\ 

        Mistral-7B-SFT-VIF-RAG & \textbf{51.2} & \textbf{56.5} & \textbf{53.8} & \textbf{45.9} & \underline{70.5} & \textbf{58.2} & \textbf{44.9} & \underline{43.0} & \textbf{44.0} & \textbf{47.8} & 58.0 & \textbf{52.9} & \textbf{47.4} & \underline{57.0} & \textbf{52.2} \\ 
        Deita-7B-V1.0-SFT & 31.4 & 31.5 & 31.4 & 29.0 & 42.5 & 35.8 & 26.5 & 30.5 & 28.5 & 26.3 & 40.0 & 33.2 & 28.3 & 36.1 & 32.2 \\ 
        \midrule
        Qwen1.5-7B-base & \underline{27.7} & 34.4 & 31.0 & \underline{27.7} & 45.9 & 36.8 & \underline{27.5} & 19.8 & 23.6 & \underline{29.9} & 45.8 & \underline{37.9} & \underline{28.2} & 36.5 & 32.3 \\ 
        Qwen1.5-7B-SFT & 16.1 & \textbf{50.5} & \underline{33.3} & 14.3 & \underline{70.0} & \underline{42.2} & 14.8 & \underline{40.0} & \underline{27.4} & 13.7 & \underline{59.0} & 36.3 & 14.7 & \underline{54.9} & \underline{34.8} \\ 
        Qwen1.5-7B-SFT-VIF-RAG & \textbf{38.9} & \underline{41.5} & \textbf{40.2} & \textbf{35.8} & \textbf{78.0} & \textbf{56.9} & \textbf{38.1} & \textbf{45.0} & \textbf{41.6} & \textbf{31.9} & \textbf{60.0} & \textbf{45.9} & \textbf{36.2} & \textbf{56.1} & \textbf{46.2} \\ 
         \midrule
        Qwen1.5-14B-base & \underline{33.7} & 38.1 & 35.9 & \underline{32.5} & 54.7 & \underline{43.6} & \underline{32.4} & 26.5 & 29.5 & \underline{33.0} & 48.3 & 40.7 & \underline{32.0} & 41.9 & 36.9 \\ 
       
        Qwen1.5-14B-SFT & 22.0 & \textbf{54.5} & \underline{38.3} & 18.7 & \underline{66.0} & 42.3 & 18.8 & \textbf{41.0} & \underline{29.9} & 19.9 & \underline{63.0} & \underline{41.4} & 19.8 & \underline{56.1} & \underline{38.0} \\ 
        Qwen1.5-14B-SFT-VIF-RAG & \textbf{42.1} & \underline{53.0} & \textbf{47.6} & \textbf{40.1} & \textbf{71.0} & \textbf{55.5} & \textbf{38.8} & \underline{39.5} & \textbf{39.2} & \textbf{35.7} & \textbf{69.0} & \textbf{52.3} & \textbf{39.2} & \textbf{58.1} & \textbf{48.6} \\ 
    
        \bottomrule
    \end{tabular}}
    \vspace{-0.5em}
    \caption{The main results on FollowRAG.``AVG'' represents the weighted average of the corresponding IF and RAG scores. The top two results in each column are highlighted in \textbf{bold} and \underline{underlined}.}
    \label{tab:1}
\end{table*}

% \makecell[c]{\textbf{FollowBench\\(SSR)}}

\begin{table*}[!t]
    \centering
    \resizebox{\textwidth}{!}{%
    \begin{tabular}{lccccccccccc}
    \toprule
        \multirow{2}{*}[-0.5ex]{\textbf{Model}} &\multicolumn{4}{c}{\textbf{IFEval}}& \textbf{\multirow{2}{*}[-0.5ex]{\makecell[c]{FollowBench\\(SSR Avg.)}}} & \multirow{2}{*}[-0.5ex]{\textbf{MT-Bench}} & \multirow{2}{*}[-0.5ex]{\textbf{Arena-Hard}} & \multirow{2}{*}[-0.5ex]{\textbf{C-Eval}} & \multirow{2}{*}[-0.5ex]{\textbf{MMLU}} & \multirow{2}{*}[-0.5ex]{\textbf{GSM8k}} & \textbf{\multirow{2}{*}[-0.5ex]{\makecell[c]{HumanEval\\(Pass@1)}}} \\
        \cmidrule(lr){2-5}
        ~ & Pr (S) & Pr. (L) & Ins. (S) & Ins. (L) & ~ & ~ & ~ & ~ & ~ & ~ & ~ \\ \midrule
        Llama3-8B-base & 24.6 & 26.1 & 38.1 & 39.7 & 11.6 & 4.0 & 0.5 & 24.2 & 38.8 & 0.5 & 0.6 \\ 
        Llama3-8B-SFT& \underline{32.5} & \underline{34.3} & \underline{43.3} & \underline{45.4} & \underline{33.6} & \underline{5.6} & \underline{2.2} & \underline{35.6} & \underline{45.2} & \underline{12.6} & \underline{3.6} \\
        Llama3-8B-SFT-VIF-RAG & \textbf{37.0} & \textbf{42.7} & \textbf{48.8} & \textbf{54.2} & \textbf{49.2} & \textbf{6.2} & \textbf{3.2} & \textbf{39.6} & \textbf{49.6} & \textbf{22.9} & \textbf{8.0} \\ \midrule
        Mistral-7B-base & 14.6 & 15.3 & 25.8 & 27.0 & 38.0 & 3.5 & 0.6 & 31.8 & \underline{44.5} & \textbf{16.0} & \underline{25.6} \\
        Mistral-7B-SFT & \underline{23.3} & \underline{24.6} & \underline{38.4} & \underline{45.7} & \underline{42.9} & \underline{6.2} & \underline{3.1} & \underline{26.2} & 32.1 & \underline{7.3} & 13.9 \\
        Mistral-7B-SFT-VIF-RAG & \textbf{34.6} & \textbf{41.0} & \textbf{46.3} & \textbf{52.0} & \textbf{53.4} & \textbf{6.5} & \textbf{3.6} & \textbf{33.0} & \textbf{49.6} & \textbf{16.0} & \textbf{32.9} \\ \midrule
        Qwen1.5-7B-base & 25.1 & 27.9 & 37.8 & 40.6 & 38.7 & 5.4 & \underline{3.2} & \underline{72.8} & \underline{58.3} & \underline{50.6} & 36.0 \\
        Qwen1.5-7B-SFT & \underline{36.4} & \underline{39.3} & \underline{46.4} & \underline{49.4} & \underline{46.3} & \underline{5.7} & 2.1 & 69.1 & 55.5 & 48.6 & \underline{39.0} \\
        Qwen1.5-7B-SFT-VIF-RAG & \textbf{42.3} & \textbf{46.0} & \textbf{53.5} & \textbf{57.1} & \textbf{51.1} & \textbf{6.1} & \textbf{3.9} & \textbf{75.6} & \textbf{61.2} & \textbf{61.4} & \textbf{44.5} \\ \midrule
        Qwen1.5-14B-base & 35.5 & 39.0 & 46.7 & 50.2 & 45.5 & 5.8 & 6.4 & \underline{77.8} & \underline{64.7} & \underline{71.8} & \textbf{59.1} \\
        Qwen1.5-14B-SFT & \underline{38.4} & \underline{41.7} & \underline{49.4} & \underline{52.6} & \underline{49.8} & \underline{6.0} & \underline{6.5} & 76.2 & 62.0 & 71.5 & \underline{58.5} \\
        Qwen1.5-14B-SFT-VIF-RAG & \textbf{46.3} & \textbf{49.9} & \textbf{60.0} & \textbf{62.2} & \textbf{56.3} & \textbf{7.3} & \textbf{7.0} & \textbf{79.5} & \textbf{66.5} & \textbf{73.8} & \textbf{59.1} \\ \bottomrule
    \end{tabular}}
    \vspace{-0.5em}
    \caption{The cross-domain validation on 4 general instruction-following (Left 4) and 4 foundational abilities (Right 4) benchmarks.
Pr. and Ins. refer to the prompt level and instruction level metric, respectively. S or L denote the strict or loose metrics used in IFEval.}
\label{table2}
\vspace{-1em}
\end{table*}

\subsection{6.1. Experimental Setup}
\textbf{Datasets.} In this section, we evaluate over 10+ benchmarks to comprehensively evaluate the VIF-RAG. 
% achieve the instruction-following alignment of RAG systems while significantly maintaining LLM‘s inherent ability. 
For the instruction-following tasks in RAG scenarios, we use the \textbf{FollowRAG} benchmark as mentioned in Section 5, which covering 4 question-answering (QA) datasets. 
% As mentioned, FollowRAG benefits from its excellent pipeline, seamlessly adapting to diverse QA data. Consequently, we included 4 question-answering (QA) datasets across 3 types: 1) Open-Domain QA: \textbf{Natural Questions (NQ)}~\citep{NQ} and \textbf{TriviaQA (TQA)}~\citep{TriviaQA}; 2) Multi-Hop QA: \textbf{HotpotQA (HQA)}~\citep{HotpotQA}; and 3) Knowledge Base QA: \textbf{WebQuestionsSP (WebQSP)}~\citep{webqsp}.
For general instruction-following evaluation, we selected two commonly used complex instruction-following datasets, \textbf{IFEval}~\citep{zhou2023instructionfollowing} and \textbf{FollowBench}~\citep{jiang2024followbench}, along with the natural instruction dataset \textbf{MT-Bench}~\citep{zheng2024judging} and the challenging ChatBot instruction-following bench, \textbf{Arena-Hard}~\citep{li2024crowdsourced}. Additionally, to measure that the foundational abilities of LLMs, we further evaluate two widely used LLM's general abilties evaluation sets, \textbf{C-Eval}~\citep{huang2023ceval} and \textbf{MMLU}~\citep{hendrycks2021measuring}, as well as the mathematical reasoning dataset \textbf{GSM8K}~~\citep{cobbe2021training} and the code evaluation bench \textbf{HumanEval}~~\citep{chen2021evaluating}.

% \textbf{Baselines.} For baselines, we selected Mistral-7B~\citep{jiang2023mistral}, Llama3-7B~\citep{llama3}, Qwen1.5-7B, and Qwen1.5-14B~\citep{yang2024qwen2} as our backbone models, using SFT ShareGPT and four QA training sets as strong baselines. Additionally, we introduce several strong instruction-following, including Conifer~\citep{sun2024conifer}, Evol-Instruc~\citep{xu2023wizardlm} and Deita~\citep{liu2024what}. To ensure fairness, we introduced an equal-sized RAG training set into the original synthetic data of these models. More details on the baselines and implementation details can be found in the appendix.

For baselines, we select Mistral-7B~\citep{jiang2023mistral}, Llama3-8B~\citep{llama3}, Qwen1.5-7B, and Qwen1.5-14B~\citep{yang2024qwen2} as our backbone models, fine-tuning ShareGPT and four QA training sets as SFT version. Besides, we introduce several strong IF baselines, including Conifer~\citep{sun2024conifer}, Evol-Instruct~\citep{xu2023wizardlm}, and Deita~\citep{liu2024what}. To ensure fairness, we add an equal-sized RAG training set to the original synthetic data used for these models. More details on the baselines and implementation can be found in the appendix.

\subsection{6.2. Main Result}

% Our main results are shown in Table 1. In general, our VIF-RAG significantly outperforms all baselines across four datasets in different setups, clearly highlighting the superiority of our approach.  Furthermore, we have identified the following insights:

% \textbf{1) Existing instruction-following baseline methods struggle in complex RAG scenarios.} Comparison between the different base model and SFT version in Table 1 and Table 2 shows that while SFT general data like ShareGPT  truely improves IFEval performance, it actually show a performance decline in the IF aspect of FollowRAG (e.g., NQ-IF: 25.68\%$\uparrow$20.97\% in Mistral). Moreover, several strong instruction-following baselines like Conifer also perform poorly and struggle in FollowRAG's IF (HQ-IF:26.9$\uparrow$26.45). This corroborates the issue pointed out in introduction: traditional instruction-following synthetic data may improve LLMs' native instruction-following ability but often fails to generalize in RAG scenarios and even leads to a performance decrease.

Our primary findings are presented in Table \ref{tab:1}. Overall, VIF-RAG consistently surpasses all baselines in FollowRAG across multiple configurations, highlighting the clear advantages of our method. Additionally, we have discovered several key insights:

\textbf{1) Existing IF baselines struggle in complex RAG scenarios.} Comparisons between different base models and SFT versions in Tables \ref{tab:1} \& \ref{table2} show that while SFT general data like ShareGPT improves performance on IFEval, it actually shows a performance decline in the instruction-following aspect of FollowRAG (e.g., NQ-IF: 25.7$\rightarrow$21.0 in Mistral). Moreover, several strong IF baselines, such as Conifer~\citep{conifer}, also perform poorly in FollowRAG's IF aspect (HQ-IF: 26.9$\rightarrow$26.45). This corroborates the issue highlighted in the introduction: traditional synthetic data may improve LLMs' vanilla instruction-following ability but often fails to generalize in RAG scenarios, sometimes even leading to decreased performance.

\textbf{2) VIF-RAG shows exceptional IF alignment capability across various datasets, models, and parameter sizes.} It consistently outperforms all baselines by over 10\% on average accuracy, including a 44\% improvement over Llama3-base, showcasing the significant performance advantage of our method. On four detailed QA benchmarks, VIF-RAG achieves the best results across all tested backbones. Moreover, whether using Qwen1.5-7B or Qwen1.5-14B, our method maintains a stable and significant performance increase of over 10\%. These results highlight that VIF-RAG is not only plug-and-play but also exhibits strong generalization capabilities.

\textbf{3) The RAG capability is effectively preserved.} Protecting RAG capability is a core focus of RAG systems. Compared to various SFT version baselines, our VIF-RAG significantly enhances IF capability while maintaining more stable RAG performance. This allows us to be optimistic about its potential in real-world RAG system applications.

\subsection{6.3. Cross-Domain Validation}

To explore the transferability of VIF-RAG, we conduct cross-domain validation on four natural instruction-following datasets and four foundational abilities benchmarks for LLMs in Tabel \ref{table2}. Our findings are as follows:

\textbf{1) Consistent IF alignment in both standard and RAG scenarios.} Table \ref{tab:1} shows that VIF-RAG achieves remarkable IF alignment in RAG scenarios. In Table \ref{table2}, comparing Llama3-8B SFT version, VIF-RAG demonstrates strong gains on two widely-used IF benchmarks, IFEval and FollowBench, with improvements of 8.8\% (Ins.L) and 15.5\% respectively. It also maintains stable improvement across different parameter sizes (7B \& 14B). These results confirm that VIF-RAG consistently enhances IF alignment in both RAG and standard scenarios.

% 1) Consistent instruction-following alignment in both standard and RAG scenarios. Table 1 demonstrates that VIF-RAG performs remarkably in RAG scenarios. Moreover, Table 2 also shows strong improvements on two widely-used instruction-following benchmarks, IFEval (Ins.L) and FollowBench, with gains of 8.8\% and 15.6\%, respectively. VIF-RAG also maintains stable performance across models with different parameter sizes (7B and 14B). These results confirm that VIF-RAG consistently enhances instruction-following alignment in both RAG and standard scenarios.

\begin{table}[t]
    \centering
    \scriptsize
    \renewcommand{\arraystretch}{1.1}
    \setlength{\tabcolsep}{0.8mm}{
    \begin{tabular}{lcccc}
        \toprule
        \multirow{2}{*}{\textbf{Model}} & \multicolumn{2}{c}{\textbf{FollowRAG (NQ)}} & \multicolumn{2}{c}{\textbf{IFEval}} \\ 
        \cmidrule(r){2-3}\cmidrule(l){4-5}
        ~ & IF & RAG & Ins(L) & Prompt(L)\\ \midrule
        Mistral-7B-SFT-VIF-RAG & 51.6 & 56.5 & 41.0 & 52.0 \\
        \midrule
        \textit{w/o Multiple Constraints} & 46.5 (-5.1) & 52.3 (-4.2) & 37.9 (-3.1) & 48.6 (-3.4) \\ 
        \textit{w/o Chain rule Constraints} & 49.2 (-2.4) & 53.3 (-3.2) & 39.2 (-1.8) & 49.9 (-2.1) \\ 
        \textit{w/o Executor based Verification} & 43.5 (-8.1) & 56.1 (-0.4) & 33.2 (-7.8) & 47.6 (-4.4) \\ 
        \textit{w/o Consistency Verification} & 47.6 (-4.0) & 46.2 (-10.3) & 38.4 (-2.6) & 46.5 (-5.5) \\ 
        \bottomrule
    \end{tabular}}
    \vspace{-0.5em}
    \caption{Ablation study on different designs of VIF-RAG.}
    \vspace{-2em}
    \label{tab:ablation}
\end{table}

\textbf{2) Robust General IF Transferability.} To assess general instruction-following alignment, we test VIF-RAG on challenging benchmarks Arena-Hard and MT-Bench. The results demonstrate that VIF-RAG maintains consistent alignment across various backbones, with a notable 1.3\% improvement on MT-Bench for the 14B model. This reveals significant potential for larger models in achieving better natural instruction alignment.

\textbf{3) Great Preservation of foundational Abilities.} Previous research highlights that enhancing specific capabilities often compromises others~\citep{dong2023abilities,hft}. As indicated in Table \ref{table2}, VIF-RAG effectively preserves general capabilities (MMLU, C-Eval), math reasoning (GSM8K), and coding skills (HumanEval) across different configurations, with some slight performance improvements. This preservation is largely attributed to the integration of ShareGPT data in the synthesis process, demonstrating VIF-RAG's ability to balance diverse capabilities while maintaining broad applicability.

\section{6.4. Quantitative Analysis}

\textbf{Ablation Study.} To examine the effects of various components in VIF-RAG, we conduct an ablation study in Table \ref{tab:ablation}. The term "w/o" indicates versions where specific components are removed. Our key observations are:

\begin{itemize}
\item Removing any component from VIF-RAG results in decreased performance, indicating that all components, such as the complex instruction composition strategy and quality verification design, are crucial to its effectiveness.

\item The largest performance decline in FollowRAG is observed when executor verification is removed. This underscores the critical role of automated instruction-response validation in improving synthetic data quality and confirms the advantage of using LLMs to oversee instruction-following abilities through other core skills like coding.

\item Surprisingly, the consistency verification proves beneficial in preserving RAG capabilities. It effectively filters out samples with high-level semantic conflicts between instructions and queries, reducing noise in IF tasks and maintaining RAG performance integrity.
\end{itemize}

% To further analyze the impact of document quantity on instruction-following performance in RAG scenarios, we conduct further exploration in Table x. For the baseline models (SFT verlsion), there is a notable decline in instruction-following capability with an increasing number of passages in the RAG scenario. When the number of documents in FollowRAG increases from 0 to 1, the performance of the baseline models drops sharply by over 6\%. As the number of input documents is further increased to 10, both baseline models show a significant performance drop, with Qwen-14B-SFT experiencing a decline of over 10\%. This indicates that the knowledge integration brought by retrieval-augmented techniques poses a significant challenge to the instruction-following ability of existing models.
% In contrast, VIF-RAG does not exhibit the severe performance drop (\≤3\%) seen with the SFT baseline when facing the first document. As the number of documents increases to 10, the performance scaling curve remains relatively stable, further demonstrating the robustness of VIF-RAG.
\begin{figure}[t]
  \includegraphics[width=\columnwidth]{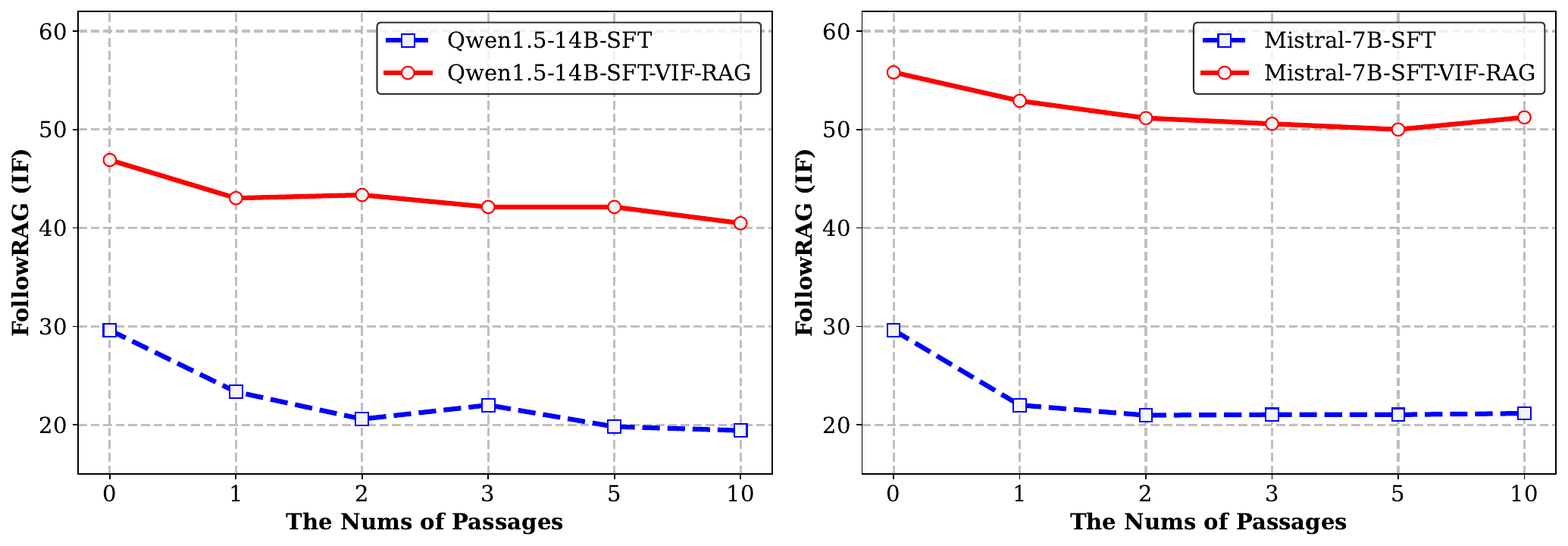}
  \vspace{-1.5em}
  \caption{The scaling analysis of retrieved document count and FollowRAG (IF) performance.}
  \label{fig:scaling}
  \vspace{-1em}
\end{figure}

\textbf{Scaling Analysis.} To explore the impact of retrieved document quantity on instruction-following performance in RAG scenarios, we refer to Table \ref{fig:scaling}. For the baseline models (SFT versions), instruction-following capability declines as the number of passages increases. Specifically, performance drops sharply by over 6\% when the document quantity in FollowRAG increases from 0 to 1. Further increasing the number to 10 leads to a significant performance decline, with Qwen-14B-SFT experiencing a drop of over 10\%. This indicates that integrating knowledge through retrieval-augmented techniques challenges the instruction-following abilities of existing models.

In contrast, VIF-RAG shows a minor performance drop ($<$3\%) when encountering the first document. As the number of documents increases to 10, VIF-RAG’s performance remains relatively stable, demonstrating its robustness.

\begin{figure}[t]
  \includegraphics[width=\columnwidth]{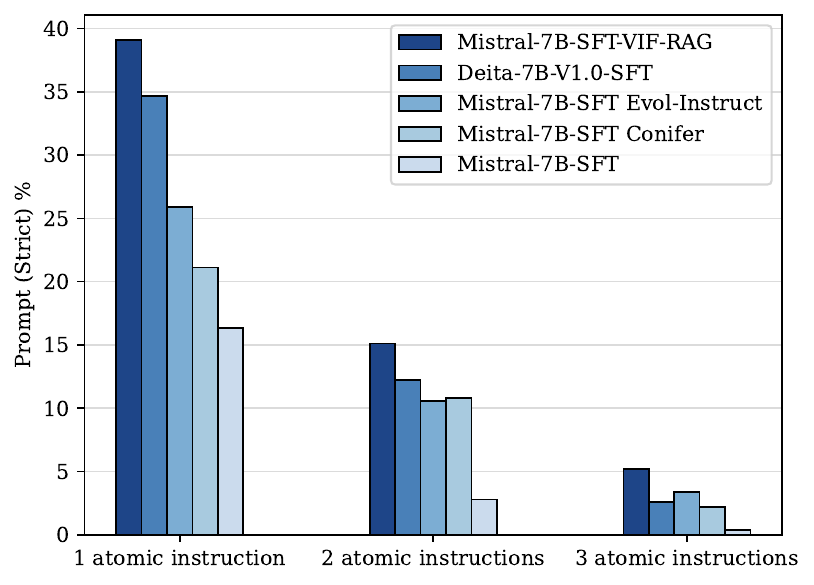}
  \vspace{-1.5em}
  \caption{The analysis of instruction counts on FollowRAG (IF) performance.}
  \label{fig:ins_num}
\end{figure}

\textbf{Instruction Difficulty Analysis.} To explore the effect of different instruction quantities (i.e., instruction-following difficulty) on model performance in RAG scenarios, we evaluate VIF-RAG and various baseline models on the FollowRAG benchmark, using test sets with 1, 2, and 3 instructions. As shown in Figure \ref{fig:ins_num}, as the number of instructions increases, all models generally show a decline in instruction-following capability, but VIF-RAG consistently outperforms the rest. Notably, even with 3 instructions present simultaneously, VIF-RAG still demonstrates over a 5\% IF prompt (strict acc.), further validating its superior capability in handling complex instruction-following tasks in RAG scenarios.

\section{7. Conclusion}
In this paper, we propose VIF-RAG, the first automated, scalable, and verifiable data synthesis pipeline for aligning complex instruction-following in RAG scenarios. VIF-RAG integrates a verification process at each step of data augmentation and combination. We begin by manually creating a minimal set of atomic instructions ($<$100) and then apply steps including instruction composition, quality verification, instruction-query combination, and dual-stage verification to generate a large-scale, high-quality VIF-RAG-QA dataset ($>$100K). To address gaps in instruction-following evaluation for RAG systems, we present FollowRAG Bench, featuring around 3K samples with 22 types of complex instruction constraints. Using FollowRAG and 8 widely-used IF and foundational abilities benchmarks, we show that VIF-RAG significantly enhances alignment on general instruction constraints and effectively demonstrates the core abilities of LLMs. Further analysis offers insights for optimizing instruction-following alignment in RAG systems.

\clearpage

\bibliography{aaai25}
\appendix

\clearpage

\section{Datasets and Baselines}
\subsection{Training Datasets}

In the data synthesis process of VIF-RAG, we use the following open-source datasets:

\begin{itemize}[leftmargin=1em]
\item \textbf{Natural Questions (NQ)}~\citep{NQ}. Natural Questions is a comprehensive dataset created to train and evaluate automatic question-answering systems. 
%它从google的网站进行大量queries的收集，并雇佣人类通过维基百科中的相关知识进行标注
It collects a large number of queries from Google's website and employs humans to annotate them using relevant knowledge from Wikipedia. The dataset contains 307,372 training examples, 7,830 for development, and 7,842 reserved for testing.

\item \textbf{TriviaQA (TQA)}~\citep{TriviaQA}. TriviaQA is an extensive and challenging text-based QA dataset comprising over 950,000 samples from 662,000 documents sourced from Wikipedia and other web pages. It is designed to test the limits of traditional QA systems, offering scenarios where answers aren't easily extracted through simple span prediction due to the lengthy and complex contexts. 

\item \textbf{HotpotQA (HQA)}~\citep{HotpotQA}. HotpotQA aims to enhance the system's ability to answer multi-hop questions and its robustness in integrating external knowledge. Unlike other QA datasets that often lack the complexity needed for training systems in reasoning and explanation, HotpotQA offers a new challenge with 113,000 question-answer pairs based on Wikipedia.

\item \textbf{WebQuestionsSP (WebQSP)}~\citep{webqsp}. WebQuestionsSP is a resource developed to assess the impact of using semantic parse labels in knowledge base question answering. It extends the original WebQuestions dataset by adding SPARQL queries for 4,737 questions and includes "partial" annotations for 1,073 questions where a complete parse was either unachievable or the questions were poorly formed or needed descriptive answers.

\item \textbf{ShareGPT}~\citep{vicuna2023}. ShareGPT is a collection of around 90,000 conversations gathered via the ShareGPT API before it was discontinued. The dataset includes user prompts and responses from OpenAI's ChatGPT, providing valuable insights into human-AI interactions. It primarily features messages in English and other Western languages, showcasing the linguistic diversity of its users.

\end{itemize}

\subsection{Evaluation Benchmarks}

For cross-domain verification in this paper, we use the following public datasets for evaluation.

\begin{itemize}[leftmargin=1em]
\item \textbf{IFEval}~\citep{zhou2023instructionfollowing}. IFEval is the most commonly used comprehensive instruction-following evaluation set for LLMs. The dataset includes over 500 prompts aimed at testing how effectively LLMs perform specific, verifiable tasks. It covers 25 types of atomic instructions, each verifiable using simple, interpretable, and deterministic programs to determine if the responses adhere to the instructions.

\item \textbf{FollowBench}~\citep{jiang2023followbench}. Followbench evaluates a model's ability to follow complex instructions by categorizing the instruction-following assessment into five different categories. It employs a multi-level mechanism to precisely enforce these constraints by evaluating the associated difficulty levels.

\item \textbf{MT-Bench}~\citep{zheng2024judging}. MT-Bench aims at evaluating multitask learning models, particularly in multi-turn dialogue and instruction-following tasks. It includes 80 high-quality multi-turn dialogue questions across eight common use cases: writing, role-playing, information extraction, reasoning, mathematics, coding, STEM knowledge, and humanities/social sciences. MT-Bench emphasizes challenging questions to effectively distinguish the capabilities of different models.

\item \textbf{Arena-Hard}~\citep{li2024crowdsourced}. Arena-Hard is a dataset used to assess the robustness of dialogue systems by testing their performance on challenging and diverse scenarios. It includes 500 carefully selected user queries that reflect complex real-world conversations, including language variations, spelling errors, and grammatical mistakes.

\item \textbf{C-Eval}~\citep{huang2023ceval}. C-Eval, as a comprehensive evaluation of general capabilities for Chinese large models, categorizes 13,948 test samples into 52 sub-domains and encompasses a difficulty system across four dimensions.. There is also a subset, C-Eval Hard, focusing on problems that demand advanced reasoning skills.

\item \textbf{MMLU}~\citep{hendrycks2021measuring}.

MMLU, as the most widely used general capability evaluation set, covers assessments across 57 domains, including science and math, with a difficulty level spanning multiple tiers. It serves as a key tool for assessing language model performance across varied tasks.

\item \textbf{GSM8K}~\citep{cobbe2021training}. GSM8K is a classic mathematical reasoning evaluation dataset that primarily focuses on math problem-solving at the grade school level. It requires models to arrive at answers by outlining their reasoning paths. The dataset contains over 8,000 samples, with the training set comprising 7,473 samples and the test set containing 1,319 samples.

\item \textbf{HumanEval}~\citep{chen2021evaluating}. HumanEval as a benchmark for evaluating code generation models, featuring 164 unique programming problems, each with about 9.6 test cases. It assesses the functional accuracy of generated code through these diverse test cases. HumanEval+ extends this by increasing the average number of test cases to 774.8 per problem, providing a more rigorous evaluation of code generation models.

\end{itemize}

\subsection{Baselines}

We compare the VIF-RAG framework with several strong baselines for the instruction-following task as follows:

\begin{itemize}[leftmargin=1em]
\item \textbf{Evol-Instruct}~\citep{xu2023wizardlm}. Evol-Instruct is the publicly available WizardLM-Evol-Instruct dataset, which includes 143k samples consisting of a blend of Alpaca and ShareGPT evolved data. In accordance with the approach outlined in the original paper. To ensure a fair comparison, we combined their dataset with the same amount of RAG data (NQ, TQ, HQ, WebQ) as used in VIF-RAG.

\item \textbf{Conifer}~\citep{sun2024conifer}. Conifer is an advanced language model designed to excel at following complex, constraint-based instructions. It stands out for its progressive learning approach, where tasks start simple and increase in complexity, allowing the model to handle intricate instructions effectively. The model's dataset was meticulously crafted using GPT-4 to ensure a diverse and challenging set of instructions. This makes Conifer particularly strong in real-world applications that require precise instruction adherence, setting it apart from other models in its ability to manage complex tasks. To ensure a fair comparison, we combined their dataset with the same amount of RAG data (NQ, TQ, HQ, WebQ) as used in VIF-RAG.

\item \textbf{Deita-7B-V1.0-SFT}~\citep{liu2023makes}. Deita-7B is a data selection method that focuses on high-quality data selection and instruction fine-tuning. It utilizes the DEITA method, which combines complexity, quality, and diversity filtering to optimize model training. Despite using a smaller dataset, Deita-7B excels in various natural language processing tasks by effectively leveraging high-quality data.

\item \textbf{SFT Version.} To build a strong baseline model and a fair comparison with VIF-RAG, we use the same amount of ShareGPT and RAG data (NQ, TQ, HQ, WebQ) as in VIF-RAG’s data synthesis process, mixing them together to fine-tune (SFT) different baseline models. This resulted in the strong baseline models labeled as "Backbone-SFT" in the main experiments.

\end{itemize}

Additionally, We list all the backbone models mentioned in the articles here.

\begin{itemize}[leftmargin=1em]
\item \textbf{Llama3-8B}~\citep{llama3}. Llama3-8B is part of the open-source Llama3 series, developed by MetaAI, is the latest and most advanced model in the Llama series. It offers notable improvements over Llama 2, Especially supporting longer input windows. These upgrades enhance performance in contextual understanding and language generation, making Llama 3 a standout in the series.

\item \textbf{Qwen1.5-7B \& 14B}~\citep{qwen}. Qwen1.5 is a significant release in Alibaba Cloud's Qwen series of large language models, featuring models with 7B and 14B parameters. It excels in multilingual tasks and supports long-context processing up to 32K tokens, making it ideal for applications like chatbots and language understanding. These enhancements made Qwen 1.5 a powerful tool for diverse AI applications.

\item \textbf{Mistral-7B}~\citep{jiang2023mistral}. Mistral 7B, released in September 2023 by Mistral AI, is an efficient language model utilizing advanced techniques. Despite having only 7 billion parameters, it outperforms many LLMs with same param size, such as reasoning, mathematics, and code generation. The model is open-source, making it widely accessible and customizable for different applications.

\end{itemize}

\section{Implementation Details}\label{app:hyperparameters}
\label{sec:prompt}

\subsection{Details about Instruction-Query Synthesis}

For the data synthesis part of RAG in section 4.2 ”Scalable Instruction-Query Synthesis“, following several RAG works~\citep{UniK-QA,dong2023bridging,luo2023chatkbqa}, we use DPR~\citep{karpukhin2020dense} as retriever for encoding knowledges. 
%w我们使用它在wikipedia检索库中根据相似度检索出topk的相关文档
We use it to retrieve the top-$K$ (k=3) relevant documents from the Wikipedia ~\citep{Wikidata} retrieval corpus based on similarity..
In this paper, we randomly samples 60K ShareGPT samples and 40K RAG samples (10K each from NQ, TQ, HQ, and WebQ), then concatenate them with our high-quality synthetic instructions.

Finally, we refer to data templates from previous RAG studies~\citep{wang2023learning,ren2023investigating,dong2024understand,qiao2024we} and directly concatenate the instruction with the RAG dataset.

% \begin{tcolorbox}[
% colback=white!10!white,
% colframe=black!75!black,
% title=Prompt Template of VIF-RAG (RAG),
% breakable]
% \{Instructions\}. Given the documents \{top-$K$ Documents\}. Answer the following question based on the given information or your internal knowledge with one or few words without the source. Query: \{Query\}. 

% \end{tcolorbox}

For the ShareGPT data, we directly concatenate the instruction with the query without using any specific templates. For consistency checks across multiple instructions in section "Instruction Composition \& Verification", The prompt are listed here:

\begin{tcolorbox}[
colback=white!10!white,
colframe=black!75!black,
title=Prompt Template for Multi-Instructions Verification,
breakable]
You are an expert proficient in determining whether multiple instructions are suitable to be implemented as simultaneous constraints.  \\

[Instructions]\textbf{\{instruction\}}\\

The text contains two or more instructions. Based on the semantic coherence and logical connection, assess whether these instructions are suitable to be implemented as simultaneous constraints. Please first conduct a thorough analysis and then assign a score ranging from 0 to 10 on the last line. A score of 0 indicates that the instructions are highly inappropriate to coexist, while a score of 10 signifies that the instructions are very suitable to serve as concurrent constraints. Please ensure that only a score is provided in the format Score: {{score}} without any additional content on the last line.

\end{tcolorbox}

% Fot the instruction-query consistency verification in section "Dual Stage Verification.", we use the following prompt template:

% \begin{tcolorbox}[
% colback=white!10!white,
% colframe=black!75!black,
% title=Prompt Template of Consistency Verification,
% breakable]
% You are an expert that is good at judging whether a response is following the instruction and query. \\
% Instruction:  \textbf{\{instruction\}} \\
% Query:  \textbf{\{query\}}  \\
% Relevant Documents:  \textbf{\{response\}}  \\
% Response:  \textbf{\{respons\}}  \\
% Please notice that the response may not be helpful as it needs to strictly follow the requirements in the Instruction. \\
% You need to judge whether the response answers the query. Please first provide a detailed analysis and then give a score ranking from 0 to 10 at the last line. \\
% Scoring 0 means the response is totally unrelated to the query, while scoring 10 means the response is helpful and highly related to the query. \\
% Please only provide a score in the format `Score: {{score}}` without any other contents at the last line.
% \end{tcolorbox}

Our VIF-RAG's prompt templates, instruction data format, verfication code, test cases and more can be found in the supplementary materials.

\subsection{Details about Supervised Fine-tuning}
% During the SFT phase, we fine-tune all 7B and 14B models with a learning rate of 7e-6, employing a linear scheduler with 20 warm-up steps. The models are trained using DeepSpeed ZeRO Stage 3~\citep{deepspeed} and Flash-Attention 2~\citep{flashattention}. Training is conducted with a global batch size of 128, a weight decay of 0.1, and runs for 3 epochs. We use mixed precision with bf16 and set a maximum context length of 4096 tokens.

%对于所有的llm微调，我们使用128的全局batchsize，输入窗口为4096，以及7e-6的学习绿配上2\%的warm up。每一组实验我们微调3个epoch。我们的训练框架为deepspeed zero 3。为了减小显存占用，我们同样采取了flash attention策略在训练，并使用bf16进行混合精读测试。

For all LLM fine-tuning, we use a global batch size of 128, an input window of 4096, and a learning rate of 7e-6 with 2\% warm-up. Each set of experiments involves fine-tuning for 3 epochs. Our training framework is DeepSpeed Zero3~\citep{deepspeed}. To reduce memory usage, we also employed the Flash Attention~\citep{dao2022flashattentionfastmemoryefficientexact} strategy during training and utilized BF16 for mixed precision testing.

Our experiments are performed on NVIDIA A800 GPUs. Specifically, Qwen1.5-7B, Mistral-7B, and LLaMA3-8B are trained on 8 A800 GPUs. We use the Llama Factory framework~\citep{zheng2024LLaMAfactory} (version 0.6.3) for training and employed greedy decoding to test the HumanEval dataset, with the metric being Pass@1. We use five sets of random seeds to conduct the same series of experiments.

%我们使用llama factory框架（版本0.6.3）进行训练，并使用greedy decoding对humaneval数据集进行测试，指标为paas@1.

\subsection{Knowledge Bases for RAG}
For the NQ, TQ, and HQ datasets, We used Wikipedia as the retrieval knowledge base. We follow the DPR approach by first applying the pre-processing code from DrQA~\citep{chen2017reading} to extract clean text, removing tables, infoboxes. Each article is then divided into 100-word text blocks, resulting in a total of 21,015,324 passages. Each passage is prefixed with the article title and an [SEP] token.

%我们使用wikipedia作为知识库

For WebQSP, we utilize Freebase~\citep{bollacker2008freebase} as the knowledge base, following the methods described in unikQA and SKP. Freebase, which includes over 125 million tuples, more than 4,000 types, and over 7,000 properties, is used for knowledge retrieval. To handle the challenge of indexing billions of relations, we implement a two-step retrieval process. DPR retrieves relations from this reduced set. The retrieved relations, typically short sentences, are combined into passages of no more than 100 tokens and provided to the FiD reader as text paragraphs.

\section{Details of FollowRAG}

\subsection{Atomic instructions in FollowRAG}
% SXS Done
We present the 22 types of atomic instructions included in FollowRAG in Table \ref{tab:atom_instruction}.

\begin{table*}[!ht]
    \centering
    \scriptsize
\renewcommand{\arraystretch}{0.9} % 增加行距，1.5表示增大为1.5倍
    % \resizebox{0.8\textwidth}{!}{%
    \setlength{\tabcolsep}{4mm}{
    \begin{tabular}{c|c|l}
    \toprule
        \textbf{Type} & \textbf{Name} & \textbf{Explanation} \\ \midrule
        \multirow{3}{*}{Keywords} & Inclusion & Include specific keywords in the response. \\ 
        ~ & Exclusion & Exclude specific keywords  in the response.\\ 
        ~ & Frequency &  Frequency constraint for including specific keywords in the response.\\ \midrule
        \multirow{3}{*}{Length} & Words &  Constraint on the number of words. \\ 
        ~ & Sentence & Constraint on the number of sentences. \\ 
        ~ & Paragraph & Constraint on the number of paragraphs. \\ \midrule
        \multirow{5}{*}{Format} & Json & Wrapped the response in JSON format. \\
        ~ & Quotation & Response wrapped in double quotes. \\ 
        ~ & No Commas & No commas allowed. \\ 
        ~ & Language & Restrict output language. \\ 
        ~ & Repeat Question & Repeat the question before answering. \\ \midrule
        \multirow{5}{*}{Structure} & Title & Include a specific title. \\ 
        ~ & Sections & Constrain the number of sections. \\ 
        ~ & Highlights & The answer must highlight at least \{N\} parts. \\ 
        ~ & Bullets & Constrain the number of bullet points. \\ 
        ~ & Placeholder & Constrain the number of placeholders. \\ \midrule
        \multirow{3}{*}{Cases} & Uppercase & Response must be in all capital letters. \\ 
        ~ & Lowercase & Response must be in all lowercase letters. \\ 
        ~ & Capital Words & Constrain the number of capitalized words \\ \midrule
        \multirow{3}{*}{Position} & End with & Response must end with specific content. \\ 
        ~ & Postscript & Use special markings at the end of the Response, such as P.S. \\ 
        ~ & First Word & Constrain the starting word of paragraph n. \\ \bottomrule
    \end{tabular}
    }
    \caption{Names and explanations of the 22 types of atomic instructions included in FollowRAG.}
    \label{tab:atom_instruction}
\end{table*}

\subsection{Judging Prompt for RAG Scores in FollowRAG}
% SXS Done

Under multiple instruction constraints, the model's target output differs from the gold answers in the original QA dataset, rendering previous evaluation metrics like exact match ineffective. To address this issue, we use the original gold answers as the reference and employ GPT-4o to assess whether the model's output correctly answers the questions.
The prompt used to instruct GPT-4o to evaluate the responses is as follows:

\begin{tcolorbox}[
colback=white!10!white,
colframe=black!75!black,
title=Judging Prompt for RAG Scores,
breakable]
% \{Instructions\}. Given the documents \{top-$K$ Documents\}. Answer the following question based on the given information or your internal knowledge with one or few words without the source. Query: \{Query\}. 
Please act as an impartial judge and perform the task: \\
Given a [Question], you need to evaluate whether the [Response] correctly answers or hits the correct answer, and output your judgment after [Judge]. I will provide a correct answer [Reference] as a reference.\\
Scoring criteria:\\
- If the [Response] is completely correct and aligns with the correct answer, it scores 1 point; \\
- If the [Response] partially answers correctly, it scores 0.5 point; \\
- If the [response] is completely incorrect compared to the [Reference], it scores 0 point.\\ \\
Note:\\
- Your only evaluation criterion is whether the [Response] correctly answered the answer, regardless of the format, language, case, length, etc., of the [Response]. Besides, providing more information than the [Reference] in the [Response] cannot be a reason for point deduction.\\
- Use the [Reference] as the correct answer reference rather than your own knowledge.\\
- The rating reply must strictly follow the format below: ``Rating: [judge\_score]\textbackslash nReason: [judge\_reason]'', and do not output any other content. For example: ``Rating: [0]\textbackslash nReason: [Response and Reference are completely unrelated.]''. Ensure that judge\_score and judge\_reason are enclosed in [].\\

[Question]\\
\{question\}\\

[Reference]\\
\{answer\_gold\}\\

[Response]\\
\{response\}\\

[Judge]
\end{tcolorbox}

Considering that evaluating the RAG scores for all samples in FollowRAG requires a substantial number of GPT-4o calls, we randomly sampled 100 entries from NQ, TQ, HQ, and WebQ for scoring and calculating the RAG scores.

\begin{table}[!t]
    \centering
    % \vspace{-1em}
    % \tiny
    \scriptsize
     \renewcommand{\arraystretch}{1.4} % 增加行距，1.5表示增大为1.5倍
    \setlength{\tabcolsep}{2mm}{
    \begin{tabular}{c|cccc}
    \toprule
        \textbf{Model} & \makecell{Qwen1.5-14B-\\base} & \makecell{Qwen1.5-14B-\\SFT} & \makecell{Qwen1.5-14B-\\SFT-VIF-RAG} & ALL \\ 
        \midrule
       \textbf{ Consistency} & 0.9639 & 0.9598 & 0.9619 & 0.9626 \\ \bottomrule
    \end{tabular}
    }
    % \vspace{-1em}
    \caption{The Pearson correlation coefficient between GPT-4o scoring and human scoring for the RAG score in FollowRAG.}

    \label{tab:consistency}
    
\end{table}

% gpt4打分和human打分的一致性
%  一共看了90条，base,sft,vifrag各30条，一共90条

%为了进一步校验我们GPT4打分的可靠性，我们进行了人工校验。

%-----------------------------------data leakage--------------------------
\begin{table}[!t]
    \centering
    % \vspace{-1em}
    \scriptsize
     \renewcommand{\arraystretch}{1.4} % 增加行距，1.5表示增大为1.5倍
    
    \setlength{\tabcolsep}{1mm}{
    \begin{tabular}{c|cccc|cc}
     \toprule
        \textbf{Setup} &\textbf{Bench.} & \textbf{Train} & \textbf{Test} & \textbf{Rephrase} & \textbf{Percentage$\downarrow$} & \textbf{N-gram$\downarrow$} \\ 
        \midrule
        
         \multirow{3}{*}{\makecell{ShareGPT+\\RAG}} & FollowRAG & 10K & 2.8K & 11 & 0.4\%& 5.3\%\\ 
         ~ & IFEval   & 10K & 542 & 2 & 0.05\%& 4.9\% \\ 
        ~ & Followbench & 10K & 820 & 1 & 0.01\%& 2.7\% \\ \bottomrule
       
        \multirow{3}{*}{VIF-RAG-QA} & FollowRAG & 10K & 2.8K & 3 & 0.1\%& 3.1\%\\ 
         ~ & IFEval   & 10K & 542 & 0 & 0.01\%& 4.3\% \\ 
        ~ & Followbench & 10K & 820 & 1 & 0.01\%& 2.6\% \\ \bottomrule

    \end{tabular}
    }
    % \vspace{-1em}
    \caption{Contamination analysis on VIF-RAG data. Train \& Test denotes the size of corresponding set. Rephr. represents samples similar to the test sample}

    \label{tab:contamination}
    
\end{table}
%-----------------------------------data leakage--------------------------

\begin{table}[h]
    \centering
    \scriptsize
    \renewcommand{\arraystretch}{1.1}
    \setlength{\tabcolsep}{2.4mm}{
    \begin{tabular}{lcccc}
        \toprule
        \multirow{2}{*}{\textbf{Model}} & \multicolumn{2}{c}{\textbf{FollowRAG}} & \multicolumn{2}{c}{\textbf{IFEval}} \\ 
        \cmidrule(lr){2-3} \cmidrule(lr){4-5}
        ~ & IF & RAG & Ins(L) & Prompt(L)\\ \midrule
        Qwen1.5-7B-base & 28.2  & 36.5  & 27.9 & 40.6 \\
        \midrule
        \textit{Supervision Model: GPT-4} & ~ & ~ & ~ & ~ \\
        % \multicolumn{5}{l}{\textit{Supervision Model: GPT-4}} \\
        Qwen1.5-7B-SFT-VIF-RAG & 36.2  & 56.1  & 46.0 & 57.1 \\
        \midrule
        \textit{Supervision Model: Qwen2-72B} & ~ & ~ & ~ & ~ \\
        % \multicolumn{5}{l}{\textit{Supervision Model: Qwen2-72B}} \\
        Qwen1.5-7B-SFT-VIF-RAG & 39.0  & 55.1  & 44.0 & 54.0 \\
        \midrule
        \multicolumn{5}{l}{\textit{Supervision Model: Llama3-70B}} \\
        Qwen1.5-7B-SFT-VIF-RAG & 40.6  & 52.3  & 41.0 & 52.3 \\
        \bottomrule
    \end{tabular}}
    \vspace{-0.5em}
    \caption{Ablation study on supervision models from GPT-4 with Qwen2-72B and Llama3-70B.}
    \label{tab:ablation_sup}
\end{table}

\subsection{Consistency with Human Evaluation}

To evaluate the effectiveness of GPT-4 scoring in assessing LLM responses, we conducted a consistency experiment between GPT-4 prediction scores and human scores. For fill-in-the-blank and open-ended questions, we randomly sampled 30 instances each from the base model, the SFT version model, and the VIF-RAG model test cases, totaling 90 instances, and had a human annotator score these predictions. In Table \ref{tab:consistency}, we report the consistency between the average human scores and GPT-4 scores, measured by Pearson correlation. The strong alignment between human and GPT-4 scores validates the effectiveness of GPT-4 scoring.

\begin{table*}[!t]
    \centering
    \resizebox{\textwidth}{!}{%
    \begin{tabular}{l|c|c|c|c|c|c|c|c|c|c|c|c|c|c|c|c|c|c|c|c}
    \toprule
        \multirow{2}{*}{\textbf{Model}} & \multicolumn{4}{c|}{\textbf{NQ}} & \multicolumn{4}{c|}{\textbf{TQ}} & \multicolumn{4}{c|}{\textbf{HQ}} & \multicolumn{4}{c|}{\textbf{WebQ}} & \multicolumn{4}{c}{\textbf{ALL}} \\ \cmidrule{2-21}
        ~ & Pr. (S) & Pr. (L) & Ins. (S) & Ins. (L) & Pr. (S) & Pr. (L) & Ins. (S) & Ins. (L) & Pr. (S) & Pr. (L) & Ins. (S) & Ins. (L) & Pr. (S) & Pr. (L) & Ins. (S) & Ins. (L) & Pr. (S) & Pr. (L) & Ins. (S) & Ins. (L) \\ \midrule
        Llama3-8B-base & 1.6 & 1.6 & 3.1 & 3.2 & 1.3 & 1.3 & 4.00 & 4.1 & 1.6 & 1.6 & 3.6 & 3.6 & 6.3 & 6.6 & 9.7 & 10.0 & 2.7 & 2.8 & 5.1 & 5.2 \\ 
        Llama3-8B-SFT & 6.1 & 6.1 & 15.7 & 15.7 & 5.3 & 5.3 & 15.0 & 15.0 & 5.1 & 5.3 & 15.0 & 15.0 & 6.1 & 6.1 & 14.4 & 14.4 & 5.7 & 5.7 & 15.0 & 15.0 \\ 
        Llama3-8B-SFT-VIF-RAG & \textbf{22.3} & \textbf{23.0} & \textbf{43.1} & \textbf{43.9} & \textbf{20.0} & \textbf{20.1} & \textbf{41.7} & \textbf{42.7} & \textbf{18.6} & \textbf{19.3} & \textbf{38.3} & \textbf{39.6} & \textbf{19.0} & \textbf{19.3} & \textbf{41.6} & \textbf{42.5} & \textbf{20.0} & \textbf{20.4} & \textbf{41.2} & \textbf{42.1} \\ \midrule
        Mistral-7B-base & 13.0 & 13.9 & 24.5 & 25.7 & 15.4 & 16.3 & 24.9 & 25.9 & 15.1 & 15.6 & 25.9 & 26.9 & 13.1 & 13.4 & 24.3 & 24.7 & 14.2 & 14.8 & 24.9 & 25.8 \\ 
        Deita-7B-V1.0-SFT & 17.1 & 18.7 & 29.3 & 31.4 & 13.7 & 14.4 & 27.0 & 29.0 & 15.3 & 16.3 & 24.8 & 26.5 & 16.6 & 16.9 & 25.1 & 26.3 & 15.7 & 16.6 & 26.6 & 28.3 \\
        Mistral-7B-SFT Conifer & 11.4 & 14.1 & 25.0 & 29.9 & 10.3 & 12.3 & 26.6 & 30.5 & 8.4 & 10.3 & 22.5 & 26.5 & 13.0 & 15.4 & 27.4 & 31.1 & 10.8 & 13.1 & 25.4 & 29.5 \\ 
        Mistral-7B-SFT Evol-Instruct & 12.7 & 21.0 & 30.7 & 41.7 & 12.3 & 17.0 & 28.0 & 37.0 & 11.0 & 15.1 & 27.7 & 35.4 & 14.1 & 18.9 & 31.8 & 39.4 & 12.5 & 18.0 & 29.5 & 38.4 \\
        Mistral-7B-SFT & 6.7 & 8.9 & 15.9 & 21.0 & 5.7 & 6.6 & 14.9 & 17.2 & 5.3 & 6.3 & 15.7 & 17.6 & 7.1 & 8.1 & 18.2 & 21.7 & 6.2 & 7.5 & 16.2 & 19.3 \\
        Mistral-7B-SFT-VIF-RAG & \textbf{19.4} & \textbf{31.3} & \textbf{37.6} & \textbf{51.2} & \textbf{17.0} & \textbf{25.6} & \textbf{34.4} & \textbf{45.9} & \textbf{17.0} & \textbf{23.7} & \textbf{33.9} & \textbf{44.9} & \textbf{20.6} & \textbf{27.9} & \textbf{37.9} & \textbf{47.8} & \textbf{18.5} & \textbf{27.1} & \textbf{35.9} & \textbf{47.4} \\ \midrule
        Qwen1.5-7B-base & 13.7 & 13.9 & 26.9 & 27.7 & 13.6 & 14.0 & 26.6 & 27.7 & 13.6 & 14.4 & 26.6 & 27.5 & \textbf{16.6} & \textbf{17.4} & 28.3 & 30.0 & 14.4 & 14.9 & 27.1 & 28.2 \\ 
        Qwen1.5-7B-SFT & 6.1 & 6.1 & 16.0 & 16.1 & 5.4 & 5.4 & 14.3 & 14.3 & 4.9 & 4.9 & 14.8 & 14.8 & 6.1 & 6.1 & 13.6 & 13.7 & 5.6 & 5.6 & 14.7 & 14.7 \\ 
        Qwen1.5-7B-SFT-VIF-RAG & \textbf{20.3} & \textbf{20.9} & \textbf{37.3} & \textbf{38.9} & \textbf{18.7} & \textbf{18.6} & \textbf{34.5} & \textbf{35.8} & \textbf{19.3} & \textbf{20.0} & \textbf{36.5} & \textbf{38.1} & 15.1 & 15.9 & \textbf{30.5} & \textbf{31.9} & \textbf{18.3} & \textbf{18.9} & \textbf{34.7} & \textbf{36.2} \\ \midrule
        Qwen1.5-14B-base & 17.6 & 18.3 & 32.7 & 33.7 & 17.7 & 18.1 & 31.8 & 32.5 & 15.7 & 16.4 & 31.6 & 32.4 & 17.1 & 17.6 & 31.7 & 33.0 & 17.0 & 17.6 & 32.0 & 32.9 \\ 
        Qwen1.5-14B-SFT & 9.6 & 9.7 & 21.7 & 22.0 & 7.4 & 7.4 & 18.3 & 18.7 & 7.4 & 7.6 & 18.7 & 18.8 & 9.0 & 9.0 & 19.5 & 19.5 & 8.3 & 8.4 & 19.6 & 19.8 \\ 
        Qwen1.5-14B-SFT-VIF-RAG & \textbf{23.6} & \textbf{24.9} & \textbf{40.7} & \textbf{42.1} & \textbf{22.3} & \textbf{22.7} & \textbf{38.5} & \textbf{40.1} & \textbf{22.9} & \textbf{23.6} & \textbf{37.3} & \textbf{38.8} & \textbf{17.9} & \textbf{18.4} & \textbf{34.8} & \textbf{35.7} & \textbf{21.7} & \textbf{22.4} & \textbf{37.8} & \textbf{39.2} \\ \bottomrule
    \end{tabular}}
    \caption{Detailed scores of instruction following under different metrics for FollowRAG. ``Pr.'' and ``Ins.'' represent Prompt and Instruction levels, while ``S'' and ``L'' denote Strict and Loose.}
    \label{tab:detailed_if_score}
\end{table*}

\section{More Experiments for VIF-RAG}

\subsection{Details of Main Results}
% SXS Done

Since FollowRAG adopts the code-based instruction following verification method following IFEval, its instruction following metrics can correspond to the two levels in IFEval as well:
\begin{itemize}[leftmargin=1em]
\item \textbf{Instruction:} The proportion of followed atomic instructions to the total number of atomic instructions in the entire dataset.
\item  \textbf{Prompt:} The proportion of samples where all atomic instructions are followed to the total number of samples in the entire dataset.
\end{itemize}
In addition, ``Strict'' and ``Loose'' indicate whether the response will be processed before scoring, such as removing common font modifiers, introductory phrases like ``Sure, here it is:'', and closing phrases like ``Hope it helps.''
We adopt the Loose Instruction score in main text and present all the different instruction follwoing scores in Table \ref{tab:detailed_if_score}.

\subsection{Data Contamination Analysis.}

We evaluate the contamination of VIF-RAG-QA on FollowRAG, IFEval and FollowBench. Our detailed analysis is conducted separately from two aspects: rule-based detection and model-based detection.

For rule-based detection,  we report contamination findings detected by traditional n-gram contamination algorithms. 
As shown in Table \ref{tab:contamination}, both contamination rates are lower than those of the ShareGPT+RAG dataset we used.

For model-based detection, we employ LLM contamination detectors from LM-Sys~~\citep{yang2023rethinking}, which utilize advanced chatbots to identify potentially rephrased contaminated test samples. Compared to ShareGPT+RAG dataset, Conifer shows relatively lower percentage of similar samples, which indicates an absence of data contamination. This allows us to confidently assert that there is no contamination between the self-generated training samples and the test sets.

% strict prompt
% 晓帅的通过率
% SXS analysis TODO

\subsection{Ablation for Supervision Model.}
\label{sec:ablation_sup}
Table \ref{tab:ablation} presents the results of replacing the supervision model from GPT-4 with Qwen2-72B and Llama3-70B. We observe that in the VIF-RAG framework, the stronger supervision model (GPT-4) demonstrates more effective strong-to-weak distillation alignment. However, Qwen2-72B and Llama3-70B also maintain solid performance, with accuracy consistency in IFEval loose prompts exceeding 50\%. This highlights the flexibility and robustness of our VIF-RAG framework, which can adapt well to different supervision models.

\section{Case Presentation and Analysis}
%Case 我已经贴到feishu里了

% \subsection{The Principles and Cases of Seed Data}
% The following introduces the four constraint themes of our manually crafted seed data:
% \begin{itemize}[leftmargin=1em]
% \item \textbf{Format Constraints} require the output to adhere to specific standards in terms of format, length, and structure. The content should be organized, clear, and meet the required format specifications.
% \item \textbf{Semantic Constraints} require the output's theme, language style, personality, and sentiment to align with the given instructions. The content should be semantically consistent with expectations and adhere to the specified tone or expression.
% \item \textbf{Knowledge Constraints} require the output to be accurate, comprehensive, and in-depth. The content should be informative, cover all necessary information, and maintain consistency in knowledge expression.
% \item \textbf{Lexical Constraints} require the output to include specific keywords or phrases, ensuring precision and relevance in word choice. The content should meet the expected requirements in terms of vocabulary selection.

% \end{itemize}

Our VIF-RAG synthetic instruction data, code, test cases, and more can be found in the supplementary materials.

\subsection{The Case Study of VIF-RAG}

\begin{figure*}[t]
    \centering
    \includegraphics[width=0.8\linewidth]{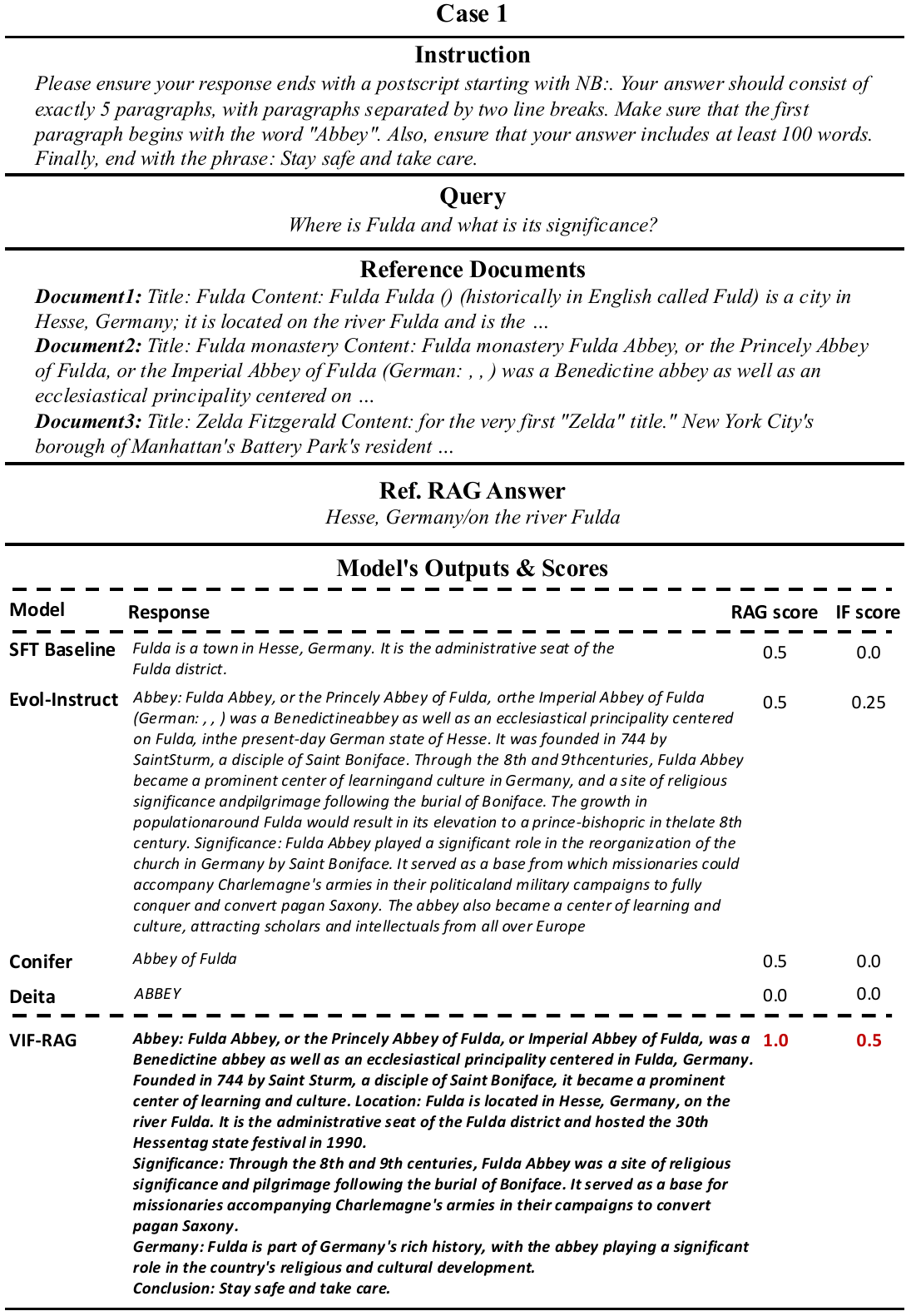}
    % \vspace{-1em}
    \vspace{-0.5em}
    \caption{A case study (1) of VIF-RAG and different models in the followrag benchmark, with the backbone Mistral-7B.}
    \label{fig:case1}
    \vspace{-1em}
\end{figure*}

\begin{figure*}[t]
    \centering
    \includegraphics[width=0.8\linewidth]{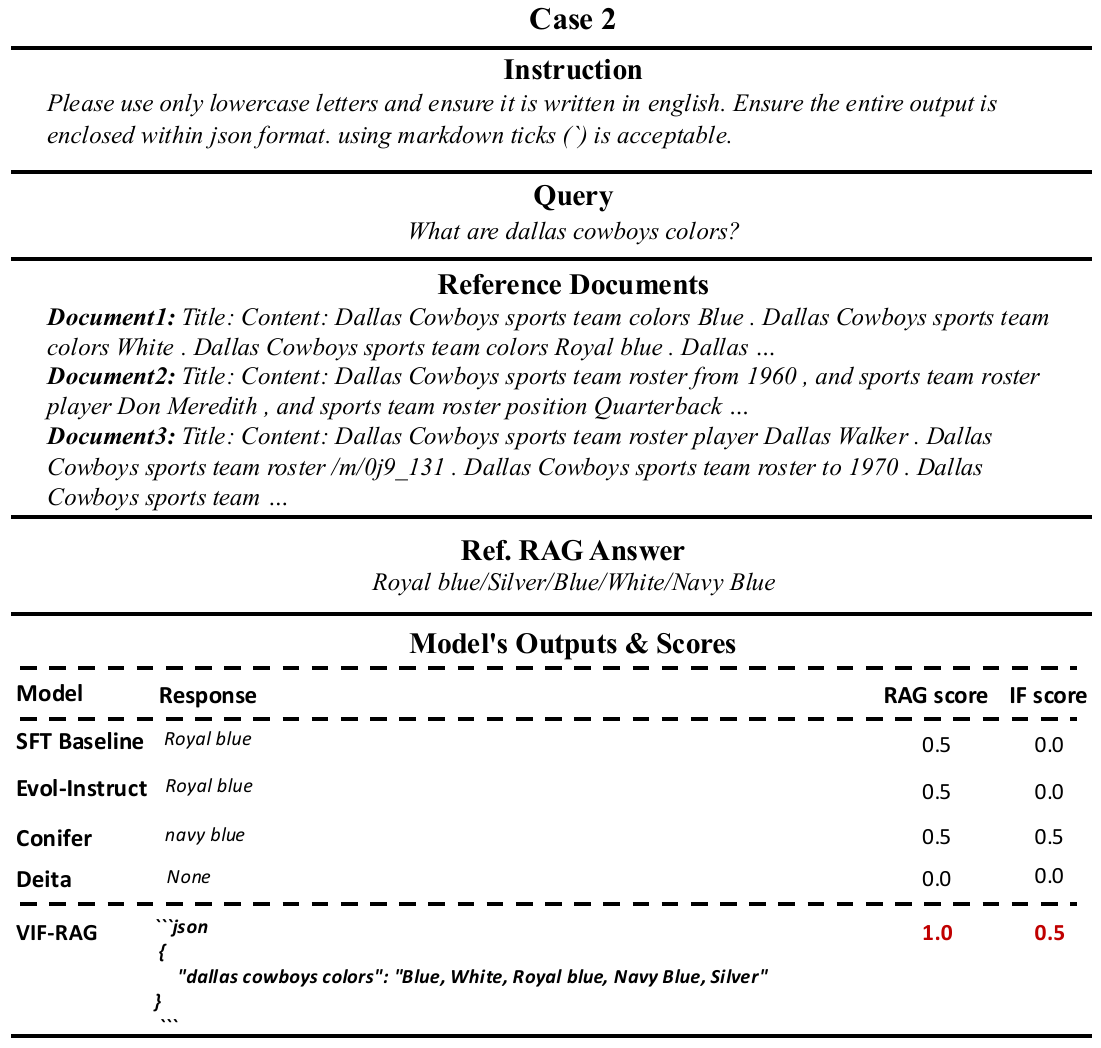}
    % \vspace{-1em}
    \vspace{-0.5em}
    \caption{A case study (2) of VIF-RAG and different models in the followrag benchmark, with the backbone Mistral-7B.}
    \label{fig:case2}
    \vspace{-1em}
\end{figure*}

\begin{figure*}[t]
    \centering
    \includegraphics[width=0.8\linewidth]{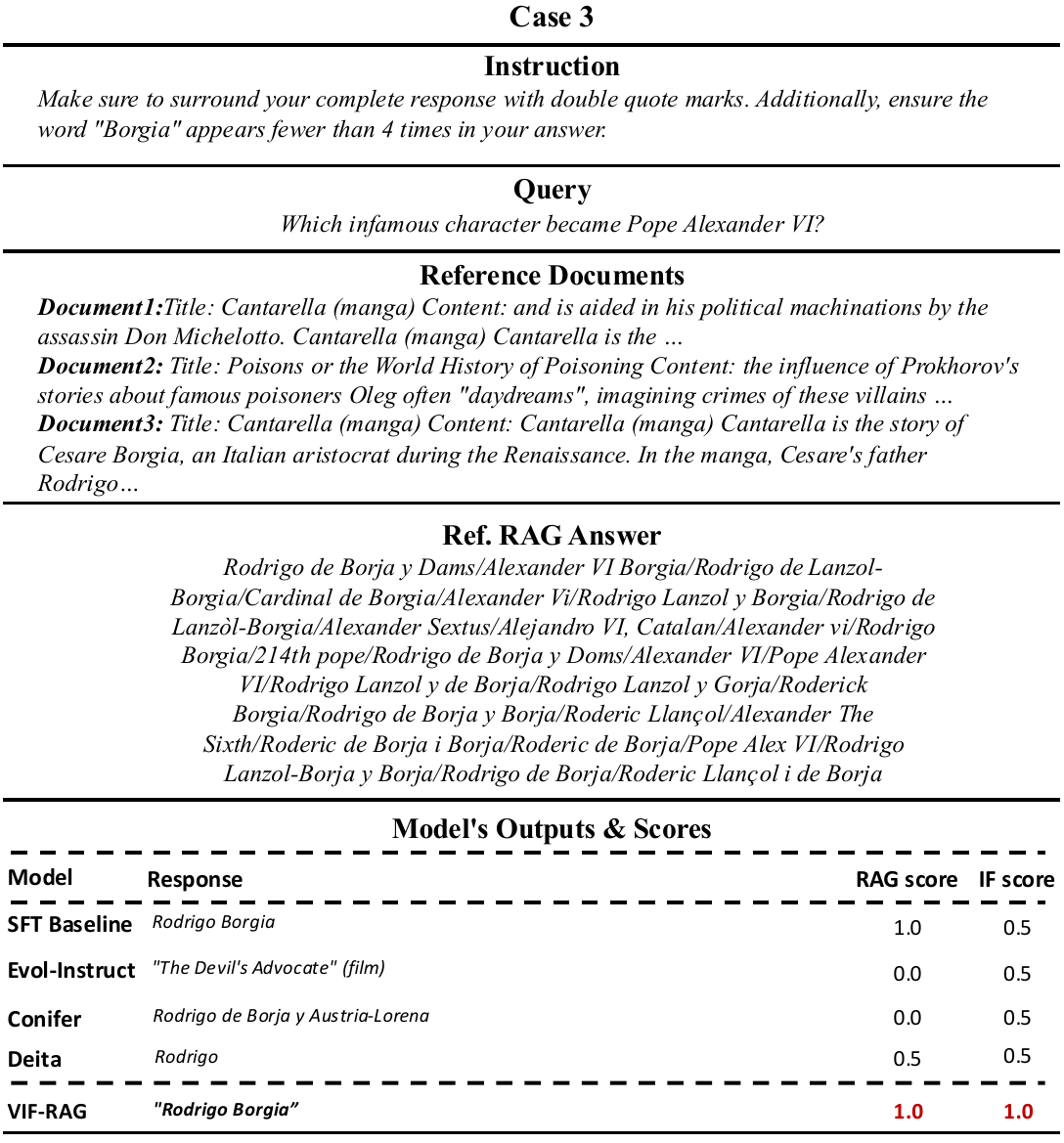}
    % \vspace{-1em}
    \vspace{-0.5em}
    \caption{A case study (3) of VIF-RAG and different models in the followrag benchmark, with the backbone Mistral-7B.}
    \label{fig:case3}
    \vspace{-1em}
\end{figure*}

To gain a deeper understanding of how VIF-RAG achieves instruction-following alignment in RAG scenarios, we conducted a case study and manual analysis, as shown in the figure \ref{fig:case1}, \ref{fig:case2} and \ref{fig:case3}. 

Since there is no truly "gold response" after following the instructions, we can only use the original gold response from the RAG dataset as a reference.

\section{Challenges and Future Work}

In this paper, we first explore instruction-following alignment in RAG scenarios and develope a high-quality RAG instruction-following data synthesis framework, VIF-RAG, along with a comprehensive benchmark, FollowRAG. However, during our research, we encounter several more challenging scenarios:

\textbf{Increased Number of Instructions:} As shown in Figure 1, our experiments revealed that existing models can handle up to 4 instructions in RAG scenarios. Even in such cases, VIF-RAG still manages to correctly answer several questions, while other baseline models lose accuracy entirely. Therefore, the challenge of increasing the number of instructions remains significant. Effectively addressing the multi-instruction problem in RAG scenarios continues to be a promising direction with major implications for complex RAG interactions.

\textbf{More Complex Instruction Types:} As the first benchmark for RAG scenarios, FollowRAG provides a comprehensive evaluation of existing models' accuracy in these scenarios. However, the variety of instructions in the real world is vast, and it is impossible to cover all types in one work. Further evaluating and improving the handling of complex instructions that are difficult to validate (e.g., interaction styles, domain-specific knowledge in RAG) will be an important focus for our future research.

We believe that future work on instruction-following alignment will offer greater promise for the practical applications of RAG systems.

\end{document}